Field Report

# Kilometer-scale autonomous navigation in subarctic forests: challenges and lessons learned

**Dominic Baril, Simon-Pierre Deschênes, Olivier Gamache, Maxime Vaidis,
Damien LaRocque, Johann Laconte, Vladimír Kubelka, Philippe Giguère** and
**François Pomerleau**
Norlab, Université Laval, Québec, QC, Canada G1V 0A6

**Abstract:** Challenges inherent to autonomous wintertime navigation in forests include lack of a reliable Global Navigation Satellite System (GNSS) signal, low feature contrast, high illumination variations, and changing environment. This type of off-road environment is an extreme case of situations autonomous cars could encounter in northern regions. Thus, it is important to understand the impact of this harsh environment on autonomous navigation systems. To this end, we present a field report analyzing teach-and-repeat navigation in a subarctic forest while subject to fluctuating weather, including light and heavy snow, rain, and drizzle. First, we describe the system, which relies on point cloud registration to localize a mobile robot through a boreal forest, while simultaneously building a map. We experimentally evaluate this system in over 18.8 km of autonomous navigation in the teach-and-repeat mode. Over 14 repeat runs, only four manual interventions were required, three of which were due to localization failure and another one caused by battery power outage. We show that dense vegetation perturbs the GNSS signal, rendering it unsuitable for navigation in forest trails. Furthermore, we highlight the increased uncertainty related to localizing using point cloud registration in forest trails. We demonstrate that it is not snow precipitation, but snow accumulation, that affects our system's ability to localize within the environment. Finally, we expose some challenges and lessons learned from our field campaign to support better experimental work in winter conditions. Our dataset is available online.[1]

## 1. Introduction

Autonomous navigation has enabled mobile robots to be deployed in a wide variety of areas in order to support human operation. Specific examples include forestry (Oliveira et al., 2021), mining (Marshall et al., 2016), disaster search and rescue (Kruijff et al., 2014), and military applications (Simon, 2015). Recently, Van Brummelen et al. (2018) have released a comprehensive review of the state-of-the-art perception technologies for autonomous vehicles. In this review, the authors identified key future challenges that need to be addressed for safer systems. One of these challenges is increasing the reliability of simultaneous localization and mapping (SLAM) algorithms to external factors such as dynamic environments, poor lighting, and weather conditions. In this regard, enabling true long-term autonomy for mobile robots will eventually require systems to be resilient to any road and weather condition.

---

[1] https://github.com/norlab-ulaval/Norlab_wiki/wiki/Kilometer-scale-autonomous-navigation-in-subarctic-forests:-challenges-and-lessons-learned

To progress towards solving the challenges mentioned by Van Brummelen et al. (2018), this research study aims to examine the impact of subarctic environments and weather on the state-of-the-art autonomous navigation approaches. Access to the Montmorency boreal forest, located 70 km north of Québec City, Canada, during winter enabled us to deploy an autonomous system in difficult meteorological conditions, which allowed us to conduct this study. Thus, we present a field report on the deployment of an autonomous navigation framework in a boreal forest under harsh winter conditions. Subarctic regions are mostly covered by the boreal forest biome, and are thus ideal for this study. Boreal forests are characterized by dense, closed-crown conifer vegetation (Russell and Ritchie, 1988) and harsh winter weather. In the forest, we distinguish two path types: forest roads and forest trails, both shown in Figure 1. Forest roads are built to accommodate various vehicles, while forest trails are narrow and are not built to accommodate typical road vehicles. The dense vegetation surrounding forest trails is not traversable by most unmanned ground vehicles (UGVs), and thus autonomous navigation error tolerance is low when navigating on these trails. Moreover, dense vegetation is known to cause problems for autonomous navigation due to interference of the canopy with the global navigation satellite system (GNSS) signal (Kubelka et al., 2020). In addition to explaining the path types, Figure 1 also demonstrates the meteorological conditions of this field deployment. It was conducted over multiple days on a 0.7-m compacted snow cover, with varying precipitation and in subzero temperatures. These conditions complicate the logistics of the deployment, diminish the endurance both of the robotic system and the personnel, and require punctual planning. Yet, they reveal weaknesses of the contemporary robotic technology and point to new research problems.

To generate observations on the impact of subarctic environments on autonomous navigation technologies, we have built the Weather-Invariant Lidar-based Navigation (WILN) system. This

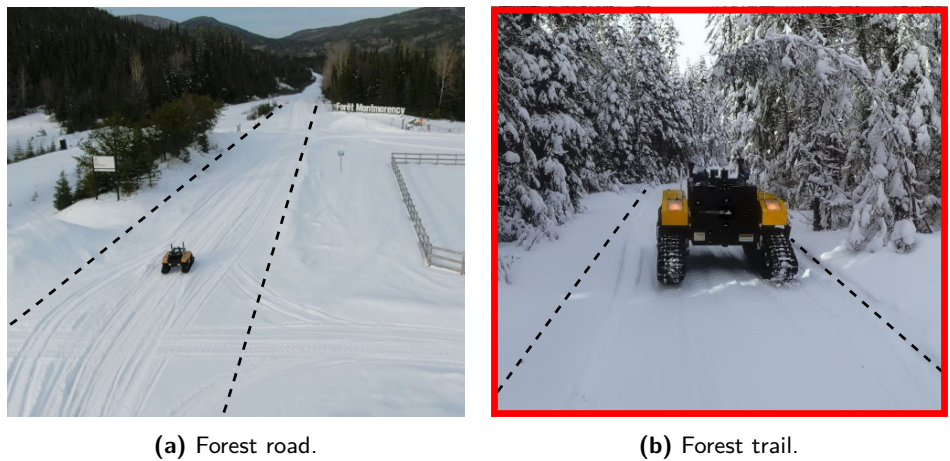

**(a)** Forest road.      **(b)** Forest trail.

**Figure 1.** The main focus of this work is evaluating the impact of biome and weather on autonomous navigation. We distinguish two distinct path types, namely forest roads and forest trails. (a) The system navigating on a wide forest path, allowing greater system inaccuracies. (b) A forest trail, where the dense vegetation prevents the robot from navigating outside the trail and blocks the GNSS signal. The error tolerance for forest trail navigation is much lower than for forest roads. In this work, the majority of autonomous navigation is conducted in forest trails, as highlighted by the red mark on the right picture.

system is a minimal autonomous teach-and-repeat framework relying primarily on lidar sensor range measurements and point cloud registration for localization. Teach-and-repeat systems require a human operator to manually teach reference paths previously to repeating them autonomously. Typical 2D localization approaches eventually fail in outdoor, three-dimensional terrain such as boreal forest roads and trails (Krüsi et al., 2015). Sensor noise due to floating particles is also known to have a strong impact on 2D localization reliability (Ren et al., 2021). Thus, the WILN system relies on 3D lidar scans to localize. Kilometer-scale environments are inherently challenging for lidar-based localization, especially due to dynamically changing environments, lack of geometrical constraints, and high computation cost related to registering point clouds within large environments. The WILN system is specifically designed and tuned to overcome such challenges and enable us to gather observations on autonomous navigation in boreal forests.

The specific contributions of this paper are (i) a comprehensive study of the impact of the boreal forest biome on lidar- and GNSS-based localization and autonomous navigation; (ii) an overview of the impact of snow accumulation on the reliability of lidar-based localization over multiple days; and (iii) a description of the WILN system, designed to enable wintertime autonomous navigation in a boreal forest. The remainder of this paper is organized as follows: Section 2 overviews the related work and robotic deployments performed in winter weather conditions. It also compares related GNSS-denied long-range autonomous navigation algorithms. Section 3 describes the WILN framework in detail, including both our localization pipeline and the path-following controller. Section 4 describes the environment in which we conducted our deployments and presents the system hardware and the implementation parameters of the WILN framework. Section 5 provides the results of the field trials and Section 6 discusses them in the context of the forest and subarctic conditions. It also presents the lessons learned during this deployment. Finally, we conclude this paper in Section 7.

## 2. Related Work

The aim of this paper is to present the impact of the boreal forest and winter conditions on autonomous navigation technologies with the goal of enabling true long-term robot autonomy. In this section, we show that while various off-road robotic deployments in winter conditions are documented in the literature, they mostly rely on the GNSS signal for localization (Lever et al., 2013). Vision-based localization approaches have enabled autonomous navigation in GNSS-denied environments. However, winter conditions have been shown to affect the performance of such approaches (Paton et al., 2017). Wintertime autonomous navigation in a boreal forest requires localization capabilities that are resilient to both winter conditions and GNSS-denied environments. Active sensors such as lidars are ideal for solving this problem since they are robust to lighting variation (Krüsi et al., 2015).

GNSS-based localization has been the standard for autonomous navigation in polar environments. The first rover to have been deployed in polar regions is *Nomad*, a gasoline-powered UGV weighing 725 kg. *Nomad* was stationed at Elephant Moraine, Antarctica, for a duration of four weeks with the goal of autonomously identifying meteorites (Apostolopoulos et al., 2000). The robot reached speeds upwards of 0.5 m s while using differential GNSS as the primary method of localization. The platform also used the stereo cameras and the lidar sensor for obstacle detection. However, stereo vision was found to be ineffective on blue ice and snow in Antarctica due to extreme lack of texture (Moorehead et al., 1999). Additionally, several operators were required in order to manually avoid undetected rocks. Expanding on this work, *MARVIN I* and *MARVIN II* were two diesel-powered skid-steering mobile robots (SSMRs) weighing 720 kg. *MARVIN I* and *MARVIN II* were deployed in Greenland (Stansbury et al., 2004) and Antarctica (Gifford et al., 2009), respectively. The goal of these robots was to increase survey safety in remote polar regions. The requirement of heavy sensor payloads led to the selection of large vehicles. *MARVIN I* used real-time kinematics (RTK) GNSS as the primary method of localization, achieving a centimeter-level accuracy in open environments. *MARVIN II* did not rely solely on RTK GNSS. When navigating out of the RTK reference receiver range, it transferred from RTK to differential GNSS. Transfer from RTK to differential GNSS increased rover oscillation with respect to the reference path. Through the *MARVIN I* and *II* deployments, Stansbury et al. (2004) highlighted the fact that turning maneuvers with SSMRs were linked to a high risk of immobilization in deep snow. Later, *Yeti*, a battery-powered 81 kg UGV, was deployed to conduct ground penetrating radar surveys in order to detect subsurface crevasses or other voids in 2010, 2011, and 2012 (Lever et al., 2013). Knowledge about such crevasses would enable human operators to plan safer paths for larger vehicles. *Yeti* did not include any obstacle detection system. During surveys, *Yeti* reached a top speed of 2.2 m/s and managed to acquire data on hundreds of crevasse encounters. It even located a previously undetected buried building in the South Pole. No obstacle detection and avoidance system was implemented on *Yeti* due to the open nature of ice sheets. Additionally, *Yeti* was deployed for further surveys in the McMurdo shear zone in 2014 (Arcone et al., 2016) and 2015 (Ray et al., 2020), while still relying on GNSS waypoint navigation. Based on the recorded data, the authors have presented a method to estimate ice sheet velocity fields by matching annual ground penetrating radar scans. The aforementioned work on field robotics in polar regions allowed the identification of various issues related to autonomous navigation on snow-covered terrain. Due to the complex nature of localizing in polar regions, GNSS is the most popular means of localization for such regions. However, subarctic navigation includes cluttered environments such as boreal forests. Autonomous navigation in those areas requires localization robust to GNSS-denied conditions, for which multiple approaches have been presented in the literature.

Since the GNSS signal is not always reliable due to the multipath effect or signal absorption due to tree canopy (Kubelka et al., 2020), approaches independent of these effects have been proposed. Furgale and Barfoot (2010) were the first to show that visual teach and repeat (VT&R) approaches are robust enough to perform large-scale autonomous navigation in GNSS-denied environments. The authors have deployed their system in the Canadian High Arctic. This environment was selected because of its similarity to lunar and Martian terrain. Most features consisted of rocks located within the reference trajectory. In this work, they successfully repeated reference paths up to 10 hours after they were manually driven. However, sensitivity to illumination change was identified as the main limit of the system. Thus, Churchill and Newman (2013) later introduced *experience-based navigation* to increase the robustness of VT&R to scene appearance change, caused by illumination variation or dynamic environment changes. This feature was added in VT&R through *multiexperience localization*, with the added ability to use landmarks from previous experiences in the same localization problem (Paton et al., 2016). In this work, the authors extend the allowable time between teach-and-repeat runs from a few hours to multiple days. Paton et al. (2015) also added color-constant image transformations to VT&R to mitigate the impact of illumination variations. Color-constant image transformations have been used by Clement et al. (2017) to perform autonomous route repeating by using the VT&R framework while relying solely on a monocular camera. While vision-based localization was demonstrated to be robust to illumination variation, Paton et al. (2015) have observed that localization frameworks relying only on passive cameras fail to localize in dark conditions.

To enable nighttime vision-based navigation, McManus et al. (2013) have proposed to use an intensity-based lidar to replace cameras for the VT&R framework. While this work demonstrates that this system is resilient to low illumination conditions, it suffers from motion distortion issues. Using headlights, MacTavish et al. (2017) proposed a bag-of-words approach to prioritize experiences most relevant to live operation. This in turn allows a growing number of robot experiences while limiting computation requirements. In this work, the authors have successfully repeated paths over a 31-h period, including day and night driving relying on headlights. Extending the experimental evaluation of VT&R, Paton et al. (2018) have logged over 140 km of autonomous navigation in an untended gravel pit, also including nighttime navigation. Clement et al. (2020) have used a deep neural network to learn a nonlinear color transform mapping that maximizes vision-based localization resiliency to appearance change. In this work, they successfully localized on routes over a 30-h period by relying only on a single experience. However, Congram and Barfoot (2021) still observed vision-based localization failures in low illumination, even when using headlights. In this work, the authors proposed to fuse vision-based localization with GNSS measurements to enable VT&R systems to function in areas where vision localization fails.

Vision-based localization was tested on snow-covered terrain numerous times. *Sno-mote Mk1* and *Mk2* were deployed on Alaskan glaciers and Wapekoneta, Ohio (Williams and Howard, 2009). *Sno-motes* are dual-drive 1:10 scale snowmobiles equipped with a single camera and GNSS. These robots were used to conduct manually driven traverses of about $100\,\text{m}$ at a speed of $1\,\text{m}\,\text{s}^{-1}$. The data gathered with the *Sno-motes* were then used to enhance visual simultaneous localization and mapping (SLAM) feature extraction methods in snow. Despite improving feature detection methods on snow, it was shown that snow is still feature sparse (Williams and Howard, 2009). Paton et al. (2017) then showed that vision-based localization is robust to intraseasonal, daily scene appearance change, by successfully repeating more than $26\,\text{km}$ over multiple days. In this work, VT&R was also deployed on a 250-m path featuring a 0.3-cm snow cover. However, deep-snow path following leads to unstable UGV behavior due to features almost only being observed on the horizon, leading to inaccurate pose estimates, which led to path-following instability. MacTavish et al. (2018) have used multiexperience localization to successfully repeat trajectories over 100 days through day, night, winter, spring, and summer. In their work, the authors showed that vision-based localization is resilient to significant seasonal change as long as the UGV can repeat the path at a rate faster than scene appearance change. In our paper, we investigate the performance of a complementary approach relying on lidar measurements and observe its limitations under harsh winter weather and in boreal forest environments.

On the other hand, lidar-based localization is resilient to illumination variation, which can lead to localization failure for vision-based systems. Marshall et al. (2008) were the first to suggest a lidar teach and repeat (LT&R) using encoders and 2D lidars. In their work, a sequence of locally consistent and overlapping topometric maps (i.e., occupancy grids) are recorded along the path using 2D lidar measurements to allow the robot to localize during the repeat phase. The system was proven efficient for repeating paths in underground tunnels on a $10\,\text{t}$ capacity hauler. Still relying only on 2D lidar scans for localization, Sprunk et al. (2013) have proposed to localize directly on 2D lidar scans, removing the requirement to build a topometric map offline. This system was deployed in an indoor, structured environment, resulting in a millimetric localization accuracy. Later, Mazuran et al. (2015) improved this framework by introducing a trajectory optimization step between the teach and the repeat phases, while still deploying the system in a similar indoor environment. Maddern et al. (2015) have studied the use of multiexperience localization, similarly to Churchill and Newman (2013), however, this time using scans measured by a push-broom 2D lidar. Local 3D swathes are produced by fusing 2D scans with vehicle odometry. These swathes are then matched to multiple prior experiences, allowing to deal with structural change in an urban environment. We show that our approach requires less memory as it relies on a single reference map and demonstrate its performance in an off-road, complex environment.

Related to off-road environments, Nieto et al. (2003) have proposed the FastSLAM algorithm relying on 2D lidar scans, which they have tested on the Victoria Park dataset. This dataset was recorded within an urban park, on uneven terrain and through sparse vegetation. In this work, the various trees in the park were used as landmarks to localize with the FastSLAM algorithm. Later, Jagbrant et al. (2015) deployed a similar lidar-based localization system in an almond orchard. However, an almond tree orchard contains significantly sparser vegetation than the boreal forest. Zhang and Singh (2018) have deployed a lidar and inertial measurement unit (IMU)-based framework in a sparse forest environment. In this work, they have shown resilience to multiseasonal change by merging a summer and winter map. However, due to the location where this work was conducted, no analysis of the impact of snowfall is discussed. Recently, Ren et al. (2021) deployed a lidar localization system in a desert biome. They identified the lack of features and geometrical constraints as an issue for point cloud registration. This issue is similar to the low feature contrast problem that affects vision-based approaches in snow-covered terrain (Paton et al., 2017).

As lidars are subject to noise created by precipitation in the environment, point cloud denoising has been studied in the literature. Schall et al. (2005) and Jenke et al. (2006) have proposed probabilistic approaches which tend to be computationally expensive. On the other hand, Schall et al. (2008) have proposed a neighborhood-based approach that is viable for real-time point cloud denoising. In a similar neighborhood-based approach, Charron et al. (2018) have proposed a dynamic radius outlier removal filter to denoise point clouds recorded by a self-driving vehicle during light snowfall in an urban setting. Duan et al. (2021) later proposed a principal-component analysis method to filter lidar scans, yielding increased performance. Utilizing the progress made on semantic segmentation, Heinzler et al. (2020) have proposed a learning-based approach to denoise point clouds, allowing to use information from the entire scene rather than the vicinity of specific points. While most point cloud denoising approaches work as input filters applied to lidar scans, we apply a post filter on the map after point cloud registration to remove dynamic points, as proposed by Pomerleau et al. (2014). We show through our results that our system is robust to real-time localization through moderate snowfall. We also show that it is not snowfall, but rather snow accumulation, that affects system performance, potentially leading to system failure.

Relying on the iterative closest point (ICP) algorithm, Krüsi et al. (2015) have deployed a LT&R system in off-road environments and busy city streets, successfully repeating paths of up to 1.3 km. These deployments have shown the resiliency of lidar-based perception to off-road and urban environments, under high illumination variations. However, this deployment does not cover the impact of dense vegetation and snowfall. In previous work, we have shown that, using the ICP algorithm to register 3D lidar scans, we can produce large-scale maps of a boreal forest offline (Babin et al., 2019). This paper aims to describe the field deployment of the WILN framework in a boreal forest during 5 days, effectively subjecting it to weak GNSS signal, high illumination variations, and low feature contrast. Such deployment conditions combine the challenges of navigation in snow-covered and GNSS-denied constrained environments. We report the impact of dense vegetation in forest trails and of snow accumulation on the performance of the WILN system.

## 3. Weather-Invariant Lidar-based Navigation (WILN) System Description

WILN is an autonomous teach-and-repeat system designed to be robust to kilometer-scale navigation, severe weather, and GNSS-denied conditions. As a localization prior, the system relies on IMU measurements and the wheel odometry, as described in Section 3.1. Registering 3D lidar scans through the ICP algorithm is the primary means of localization (Section 3.2). Map maintenance and tiling modules were added to enable kilometer-scale SLAM, both of which are described in Section 3.3. As for most teach-and-repeat frameworks, two modes are defined: the *teach phase* and the *repeat phase* (Section 3.4). A kinematic controller was selected to compute appropriate commands to solve the path-following problem (Section 3.5). This section ends with a system overview, including a visual representation of all WILN system components and their interactions (Section 3.6).

Four coordinate frames are defined in the WILN system, as illustrated in Figure 2. First, the global map frame $\mathcal{G}$ is defined, representing the world the robot navigates in. Second, the robot frame $\mathcal{R}$ is defined with its origin in the base of the robot chassis. The $x$ axis of the robot frame $\mathcal{R}$ is parallel to the longitudinal direction and the $y$ axis is parallel to the lateral direction of the robot. Third, the lidar frame $\mathcal{L}$ coincides with the lidar sensor origin. The rigid transform from the robot frame to the lidar frame $^{\mathcal{L}}_{\mathcal{R}}\boldsymbol{T}$ is assumed to be constant and found through system calibration. The reading point clouds (i.e., input points) $\mathcal{P}$ are originally observed in the lidar frame $\mathcal{L}$ and the reference point clouds (i.e., map points) are expressed in the map frame $\mathcal{G}$. The transform $^{\mathcal{G}}_{\mathcal{R}}\boldsymbol{T}$ between the frames $\mathcal{R}$ and $\mathcal{G}$ is updated by the ICP algorithm and constitutes the robot localization. For the path-following algorithm running in the repeat phase, the Frenet-Serret frame $\mathcal{S}$ is defined directly on the path. Lastly, the reference trajectory recorded in the teach phase is expressed as a vector of consecutive robot poses $\boldsymbol{x}_{\text{ref}} = \{^{\mathcal{G}}_{\mathcal{R}}\boldsymbol{T}_1, ^{\mathcal{G}}_{\mathcal{R}}\boldsymbol{T}_2, \ldots, ^{\mathcal{G}}_{\mathcal{R}}\boldsymbol{T}_n\}$, where $n$ is the number of recorded poses in the trajectory.

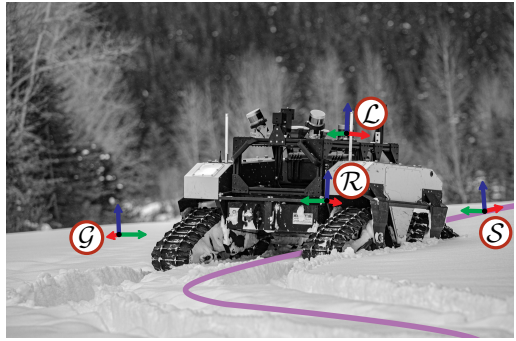

**Figure 2.** Coordinate frames used for WILN. In this instance, the robot is moving towards the path frame $\mathcal{S}$. The reference path that the robot aims to follow is drawn in light purple. The remaining relevant coordinate frames are the map frame $\mathcal{G}$, the lidar frame $\mathcal{L}$, and the robot frame $\mathcal{R}$.

### 3.1. Localization Prior: IMU and Wheel Odometry

The ICP algorithm requires high-frequency prior estimates of $^{\mathcal{G}}_{\mathcal{L}}\check{\boldsymbol{T}}$ that capture the motion between every two consecutive lidar scans. In WILN, the prior $^{\mathcal{G}}_{\mathcal{L}}\check{\boldsymbol{T}}$ is estimated through IMU measurements and wheel odometry. The robot orientation is estimated using the Madgwick filter[2] (Madgwick et al., 2011) based on gyroscope and accelerometer measurements. Linear displacement is based on wheel odometry, while taking into account the estimated robot orientation. The prior $^{\mathcal{G}}_{\mathcal{L}}\check{\boldsymbol{T}}$ is generated synchronously with the IMU at a frequency of 100 Hz. When subject to motion, it should be noted that lidar sensors generate skewed reading point clouds $^{\mathcal{L}}\mathcal{P}_s$. Hence, we deploy a point cloud deskewing filter in our system. First proposed by Bosse and Zlot (2009), such algorithms correct point cloud distortions by taking the lidar intrascan motion into account. Following the same idea, our high-frequency $^{\mathcal{G}}_{\mathcal{L}}\check{\boldsymbol{T}}$ prior is used to deskew the raw input point cloud. The resulting corrected $^{\mathcal{L}}\mathcal{P}$ is the one used in the ICP algorithm.

### 3.2. Iterative Closest Point (ICP)

To localize the robot, the incoming deskewed reading point clouds $^{\mathcal{L}}\mathcal{P}$ are registered to the reference point cloud $^{\mathcal{G}}\mathcal{Q}$ using the ICP algorithm. This allows a map of the environment to be built during the teach phase and enables the robot to be localized in this map during the repeat phase. An overview of the ICP pipeline is shown in Figure 3. The ICP algorithm iteratively matches points between two point clouds and looks for a rigid transform that minimizes the distance between each pair of the matched points. Pomerleau et al. (2015) presented a comprehensive review of the state of the art for the ICP algorithm. The WILN registration component is based on the modules presented in this review. To increase robustness of the algorithm, we apply the following three input filters to the reading $^{\mathcal{L}}\mathcal{P}$ before registering it into the reference $^{\mathcal{G}}\mathcal{Q}$:

1. **Random subsampling filter** parameterized by the ratio $\eta_s \in [0,1]$ of the points kept after the subsampling. This subsampling is critical to reducing the computation time of the ICP algorithm to allow the SLAM problem to be solved in real time (Pomerleau et al., 2011).
2. **Bounding box filter** parameterized by the bounding box coordinates $\boldsymbol{b}_i = (x_{\min}, x_{\max}, y_{\min}, y_{\max}, z_{\min}, z_{\max}) \in \mathbb{R}^6$. It removes points originating from the robot body that would otherwise cause a trail of points in the reference point cloud $\mathcal{Q}$.
3. **Radius filter** parameterized by the maximum radius $r$ around the lidar. Beyond this radius, the reading points are discarded. This allows only the relevant vicinity around the robot to be considered to further reduce the computation time.

---

[2] https://github.com/bjohnsonfl/Madgwick_Filter

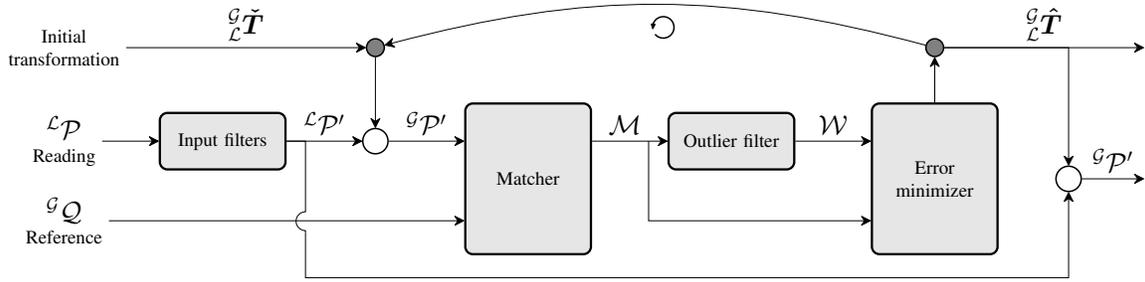

**Figure 3.** The ICP pipeline. The reading point cloud $^{\mathcal{L}}\mathcal{P}$ is filtered, and the initial transformation $^{\mathcal{G}}_{\mathcal{L}}\check{T}$ is applied to it. The matcher finds sets of neighbors in $^{\mathcal{G}}\mathcal{Q}$ for each point of $^{\mathcal{G}}\mathcal{P}'$, which compose the set of matches $\mathcal{M}$. An outlier filter is then used to compute weights $\mathcal{W}$ associated with the matches $\mathcal{M}$. The weights $\mathcal{W}$ and the matches $\mathcal{M}$ are then used in the error minimizer to compute an output transformation $^{\mathcal{G}}_{\mathcal{L}}\hat{T}$. The matching, outlier filtering, and error minimization steps are done iteratively until the error is beneath the settled threshold.

In our implementation, $^{\mathcal{L}}\mathcal{P}$ is observed in the lidar frame $\mathcal{L}$ and $^{\mathcal{G}}\mathcal{Q}$ is defined in the map frame $\mathcal{G}$. Thus, the ICP algorithm estimates the transform $^{\mathcal{G}}_{\mathcal{L}}\hat{T}$ by minimizing an error function e using

$$^{\mathcal{G}}_{\mathcal{L}}\hat{T} = \arg\min_{T} \left[ \text{e}\left(^{\mathcal{G}}\mathcal{P}, ^{\mathcal{G}}\mathcal{Q}\right) \right], \tag{1}$$

where $^{\mathcal{G}}\mathcal{P}$ is the reading point cloud $\mathcal{P}$ expressed in the map frame $\mathcal{G}$. For the first iteration, the prior $^{\mathcal{G}}_{\mathcal{L}}\check{T}$ is used to compute this rigid transformation. The prior computation is explained in Section 3.1. To compute the error function, we associate points between the reading point cloud and the reference point cloud. Following this step, the ICP algorithm computes the optimal transform by iteratively minimizing the error between $^{\mathcal{G}}\mathcal{P}$ and $^{\mathcal{G}}\mathcal{Q}$, as specified in Equation 1. For better clarity, the remainder of this section concentrates on a single ICP iteration step that would be repeated until convergence.

Point matching is first done by finding the closest points in $^{\mathcal{G}}\mathcal{Q}$ for each point of $^{\mathcal{G}}\mathcal{P}'$. Thus, multiple points of $^{\mathcal{G}}\mathcal{Q}$ can be associated with each point of $^{\mathcal{G}}\mathcal{P}'$. Nearest-neighbor search is carried out via the use of a kd-tree[5] to decrease computation time. The matcher is parameterized by the number of nearest neighbors for each $\mathcal{P}$ point $n_m \in \mathbb{Z}_{>0}$ and the maximum allowable distance for a match, $d_{\max} \in \mathbb{R}_{>0}$. In order to speed up the nearest-neighbor searches, the $\varepsilon \in \mathbb{R}_{>0}$ parameter is set to 1 to allow approximations, as described in Arya and Mount (1993). Subsequently, we apply an outlier filter to add binary weights to each matched point to remove the outlier matches from the error function. The outlier filter adds a positive binary weight to the $\eta_d \in [0,1]$ proportion of nearest matches. Formally, let $\mathcal{M}_{\text{all}} = \text{match}(\mathcal{P}, \mathcal{Q}) = \{(\boldsymbol{p}, \boldsymbol{q}) \in \mathcal{P} \times \mathcal{Q}\}$ be the set of matches between $\mathcal{P}$ and $\mathcal{Q}$. We also define $\mathcal{M} \subseteq \mathcal{M}_{\text{all}}$, which contains only the $n_m$ closest pairs of $\mathcal{M}$ for each point of $\mathcal{P}$ with a Euclidean distance below $d_{\max}$. Let $\mathcal{W} = \text{outlier}(\mathcal{P}, \mathcal{Q}) = \{\text{w}(\boldsymbol{p}, \boldsymbol{q}) : (\boldsymbol{p}, \boldsymbol{q}) \in \mathcal{M}\}$ be the weights associated with these matches. Our system uses the point-to-plane error function, defined by

$$\text{e}(\mathcal{P}, \mathcal{Q}) = \sum_{k=1}^{K} \text{w}(\boldsymbol{p}_k, \boldsymbol{q}_k) \left\| (\boldsymbol{p}_k - \boldsymbol{q}_k) \cdot \boldsymbol{n}_k \right\|_2, \tag{2}$$

where $K$ is the number of matches in $\mathcal{M}$ and $\|\cdot\|_2$ is the L2 norm. The normal vector $\boldsymbol{n}_k$ around the 3D point $\boldsymbol{q}_k$ in $\mathcal{Q}$ is computed prior to the ICP algorithm. The error in Equation 2 can be iteratively minimized by recomputing the set of matched points $\mathcal{M}$ and their associated weights w at each iteration. Moreover, in the minimization process, we only optimize the translation and the yaw rotation, assuming that the prior roll and pitch angles are already optimal. This holds in the case of precise IMU calibration with respect to the robot frame $\mathcal{R}$. Transformation checkers are added to WILN to detect erroneous ICP solutions. An error is raised if the prior error exceeds predefined thresholds in translation error $\varepsilon_{t_{\min}} \in \mathbb{R}_{>0}$ or angular error $\varepsilon_{\theta_{\min}} \in \mathbb{R}_{>0}$. Additionally, the iterative process of the ICP algorithm is stopped after a maximum number of iterations, $i_{\max} \in \mathbb{Z}_{>0}$, is reached, returning the last transform. The resulting transform $^{\mathcal{G}}_{\mathcal{L}}\hat{T}$ can be used to express the 3D robot pose in $\mathcal{G}$ by chaining $^\mathcal{G}_\mathcal{L}\hat{\boldsymbol{T}}\,^\mathcal{L}_\mathcal{R}\boldsymbol{T}$ since the latter was evaluated through system calibration. The ICP algorithm also returns the reading point cloud in the map frame $^\mathcal{G}\mathcal{P}$. Our implementation is strongly based on the `libpointmatcher` library (Pomerleau et al., 2013).[3] This library allows fast registration of large point clouds, which is essential to solving the SLAM problem in real time.

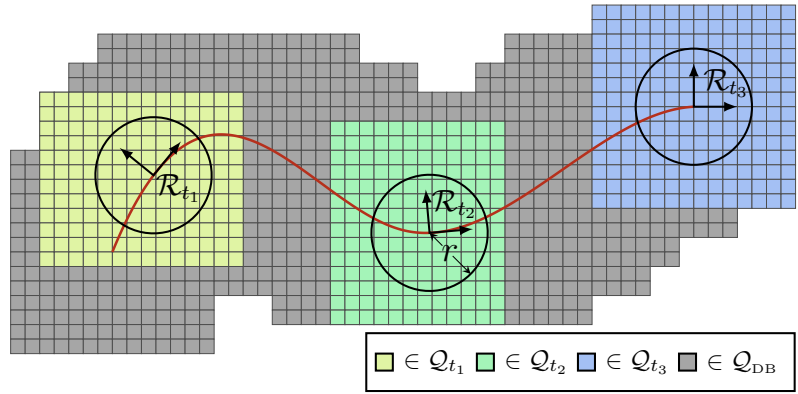

**Figure 4.** The management of the map by the *voxel manager* module. The robot trajectory is represented by the red line. Three distinct robot poses are represented by $\mathcal{R}_{t_1}$, $\mathcal{R}_{t_2}$, and $\mathcal{R}_{t_3}$. The local map $^\mathcal{G}\mathcal{Q}$ voxels are represented by the light yellow, green, and blue colors and the nonlocal map $^\mathcal{G}\mathcal{Q}_{\text{DB}}$ voxels by the dark gray color.

### 3.3. Large-Scale Mapping

The large-scale mapping subsystem uses the ICP implementation described in Section 3.2 to solve the SLAM problem in large-scale navigation. At each time step, a lidar scan $^\mathcal{L}\mathcal{P}$ is measured and registered to the reference map $^\mathcal{G}\mathcal{Q}$, yielding the transformed and filtered reading $^\mathcal{G}\mathcal{P}$ and the estimated transform between the lidar and map frames $^\mathcal{G}_\mathcal{L}\hat{\boldsymbol{T}}$. The registered scan $^\mathcal{G}\mathcal{P}$ is appended to the reference point cloud $^\mathcal{G}\mathcal{Q}$. To maintain a point density in the map, a point is only appended if its distance to the closest point in the map exceeds the user-defined minimal threshold $\rho \in \mathbb{R}_{>0}$. After merging the input point cloud into the map, two post-filters are applied to the map:

1. The **surface normal filter** computes the surface normals for all points required by the point-to-plane minimization described in Equation 2. For each map point, the positions of the $n_n \in \mathbb{Z}_{>0}$ closest points are taken into account to estimate the local surface normal.
2. The **dynamic point filter** removes points that are identified as dynamic, originating from, e.g., walking pedestrians, or falling snow. Its implementation is based on the one proposed by Pomerleau et al. (2014). The dynamic filter allows dynamic elements to be removed from the map, allowing the map to be maintained for extended periods of time. To filter out the dynamic points, ray tracing is used. If an incoming scan $^\mathcal{L}\mathcal{P}$ point is located behind a map point $^\mathcal{L}\mathcal{Q}$, the probability of this map point being dynamic is increased. If this probability surpasses the predefined threshold $\tau_d \in [0, 1]$, it is effectively removed from $^\mathcal{G}\mathcal{Q}$. To limit computation time for this filter, map points located further than the sensor range $r$ do not enter the filtering process.

This new map $^\mathcal{G}\mathcal{Q}'$ is sent to the *voxel manager* module, illustrated in Figure 4. The purpose of the voxel manager is to limit the computational complexity of the map maintenance, instead of letting it grow with its size. Our implementation of the voxel manager module is similar to the one proposed by Ren et al. (2021). It involves the following steps: the new map $^\mathcal{G}\mathcal{Q}'$ is divided into voxels, then only the voxels that are close to the robot are kept in the random access memory (RAM). All points located within those voxels are considered to be the local map $^{\mathcal{G}}\mathcal{Q}$. The remaining voxels represent the nonlocal map $^{\mathcal{G}}\mathcal{Q}_{\text{DB}}$ and are stored in the database on the system hard drive.

---

[3] https://github.com/ethz-asl/libpointmatcher

The local map $^{\mathcal{G}}\mathcal{Q}$ voxels constitute a cube following the position of the $\mathcal{L}$ frame. The center of this cube is maintained close to the origin of the $\mathcal{L}$ frame. The initial length of its edge is defined as twice the lidar sensor range $r$ plus a margin of two voxels on each side. This margin gives the manager enough time to load and unload voxels as the robot moves through the map. The loading and unloading operations are triggered each time the robot crosses two distinct voxel borders, preventing unnecessary computations if the robot oscillates on a single voxel border. The side effect is the varying number of voxels in the local map $^{\mathcal{G}}\mathcal{Q}$, as can be observed in the example for the second robot pose $\mathcal{R}_{t_2}$ in Figure 4. In this case, tile loading will be triggered once the sensor range crosses the next voxel frontier, which explains why only one voxel separates the sensor range from the local map edge. The voxel manager allows the maximum map maintenance computation time to be limited, allowing the system to solve the SLAM problem for kilometer-scale environments. This module is parametrized by the voxel size $v_s \in \mathbb{R}_{>0}$, in meters.

### 3.4. Teach and Repeat Phases

During the teach phase, the robot is driven along a specific path by a human operator. The large-scale mapping framework presented in Section 3.3 is used to build the reference map. The local $^{\mathcal{G}}\mathcal{Q}$ and nonlocal $^{\mathcal{G}}\mathcal{Q}_{\text{DB}}$ map voxels are saved to a map database. The sequence of robot poses $\boldsymbol{x}_{\text{ref}}$ estimated by the ICP algorithm is subsampled to keep only points that are at least $d_{\text{ref}}$ apart from each other. This subsampled trajectory is also saved in the database and constitutes the reference path $\boldsymbol{x}_{\text{ref}}$ for the repeat phase. The system can be taught various reference trajectories and maps, all of which are stored on the robot's hard drive.

During the repeat phase, the complete reference map and the reference path $\boldsymbol{x}_{\text{ref}}$ are first loaded from the database. The voxel manager then rebuilds the local map $^{\mathcal{G}}\mathcal{Q}$ and the nonlocal map $^{\mathcal{G}}\mathcal{Q}_{\text{DB}}$ voxels according to the robot's position. In this phase, no points are appended to the reference map and the ICP algorithm is only used to localize the robot in the local map. One should be cognizant that snow accumulation is heterogeneous in nature, meaning that it is significantly different on dense vegetation, buildings, and flat ground. Thus, dynamic map updates were not added to the repeat phase to ensure system reliability during this deployment. The estimated pose of the robot $^{\mathcal{G}}_{\mathcal{R}}\hat{\boldsymbol{T}}$ is then projected as a planar pose $\boldsymbol{x}_{\text{2D}}$ and used by the path-following algorithm described in Section 3.5. The path-following algorithm computes the output system commands $\boldsymbol{u}$ to steer the UGV along the $\boldsymbol{x}_{\text{ref}}$ poses. Note that we rely on a metrically consistent map, contrarily to VT&R, which relies on viewpoint-based localization (Furgale and Barfoot, 2010). This choice allows us to maintain a single 3D map, which is memory efficient comparatively to viewpoint-based localization. Indeed, the latter requires storing all experiences within a database, eventually leading to significant memory usage.

### 3.5. Path Following

Once the robot's pose $^{\mathcal{G}}_{\mathcal{R}}\hat{\boldsymbol{T}}$ has been estimated and the reference trajectory $\boldsymbol{x}_{\text{ref}}$ loaded from the database, this information is used as the input to the path-following controller. The outputs of the path-following algorithm are the commanded longitudinal and angular velocities, respectively defined in the vector $\boldsymbol{u} = [v_x, \omega] \in \mathbb{R}^2$. The WILN framework uses the orthogonal-exponential (OE) controller to compute the command vector $\boldsymbol{u}$. This controller computes angular velocity through the orthogonal projection of the robot pose on the reference path and various heuristics to compute linear velocity (Huskić et al., 2017). For the WILN system implementation, we have used the Generic Robot Navigation (GeRoNa)[4] library proposed by Huskić et al. (2019) to allow fast computation of the command vector $\boldsymbol{u}$.

---

[4] https://github.com/cogsys-tuebingen/gerona

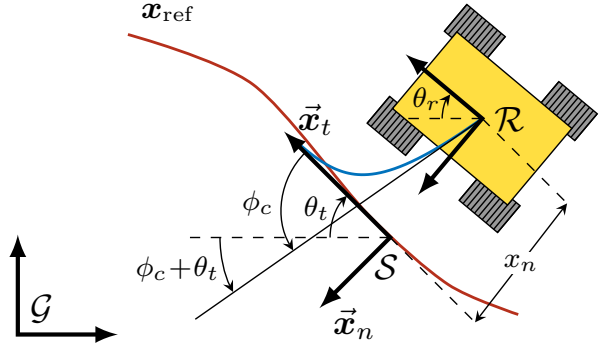

**Figure 5.** Main components of the orthogonal-exponential (OE) path-following algorithm used in this work. In red is the reference trajectory. In blue is the predicted exponential trajectory. The Frenet-Serret frame $\mathcal{S}$ is located on the referential trajectory.

At each time step, a control law is built to allow the robot to adjust its orientation to converge to the reference path $\boldsymbol{x}_{\text{ref}}$, as shown in Figure 5. A Frenet-Serret frame $\mathcal{S}$ is defined in the map frame. Its origin lies in the orthogonal projection of the robot frame $\mathcal{R}$ on the reference path. After, an exponential function is defined to allow the robot to converge to the reference path, drawn in blue in Figure 5. The angle of this exponential function is defined by

$$\phi_c = \arctan(-kx_n \exp(-kx_t)), \tag{3}$$

where $k \in \mathbb{R}_{>0}$ is a constant that allows the regulation of the convergence speed of the robot to the path. We also denote $x_t \in \mathbb{R}$ and $x_n \in \mathbb{R}$ as the current position of $\mathcal{R}$ in path frame $\mathcal{S}$, along the tangential $\vec{\boldsymbol{x}}_t$ and normal $\vec{\boldsymbol{x}}_n$ axes, respectively. Linear velocity is computed by modulating the target vehicle velocity $v_n$. Thus, the complete control law can be defined by

$$\boldsymbol{u} = \begin{bmatrix} v_x \\ \omega \end{bmatrix} = \begin{bmatrix} v_n \exp\left(-\left(\frac{K_g}{d_g}\right)\right) \\ K_h(\phi_c + \theta_e) \end{bmatrix}, \tag{4}$$

where $\theta_e \in \mathbb{R}_{>0}$ is the error between the robot angle $\theta_r$ and the path frame angle $\theta_t$ in the global frame $\mathcal{G}$. The $K_h \in \mathbb{R}_{>0}$ parameter is a gain on commanded angular velocity that was added in order to reduce controller overshoot when subject to high reference path curvature $\kappa$. In the original implementation, a parameter exists allowing to reduce commanded longitudinal velocity in areas with high path curvature $\kappa$. However, this parameter was omitted due to noise in the teach phase localization resulting in jerking motion for the UGV. The distance between the current robot pose and the end of the trajectory is represented by $d_g \in \mathbb{R}_{>0}$ and a goal proximity gain $K_g \in \mathbb{R}_{>0}$ is defined to gradually slow the robot as it reaches its goal. To respect the limitations of the robotic platform, the commanded angular velocity is limited to $\omega \in [-\omega_m, \omega_m]$. The commanded linear velocity is bounded to $v_x \in [v_{\min}, v_{\max}]$. A goal tolerance $\tau_g \in \mathbb{R}_{>0}$ parameter is used to allow the robot to finish the path repetition when within an acceptable distance from the $\boldsymbol{x}_{\text{ref}}$ end. Finally, a safety tolerance $\tau_w \in \mathbb{R}_{>0}$ parameter is defined to stop the robot if the distance between the origin of the robot's frame $\mathcal{R}$ and the closest trajectory point of $\boldsymbol{x}_{\text{ref}}$ exceeds the specified distance $\tau_w$.

### 3.6. System Overview

An overview of the entire WILN pipeline is shown in Figure 6. During the teach phase, an operator drives the robot along the desired trajectory. Sensor measurements are used to solve the SLAM problem and build a database containing all reference maps and trajectories. A point cloud deskewing system is used to take intrascan lidar motion into account. This same system is used to produce a transformation prior $^{\mathcal{G}}_{\mathcal{L}}\check{\boldsymbol{T}}$ to allow the ICP algorithm to perform real-time point cloud registration. A map maintenance module is used to append the latest registered lidar scan $^{\mathcal{G}}\mathcal{P}$ to the local map

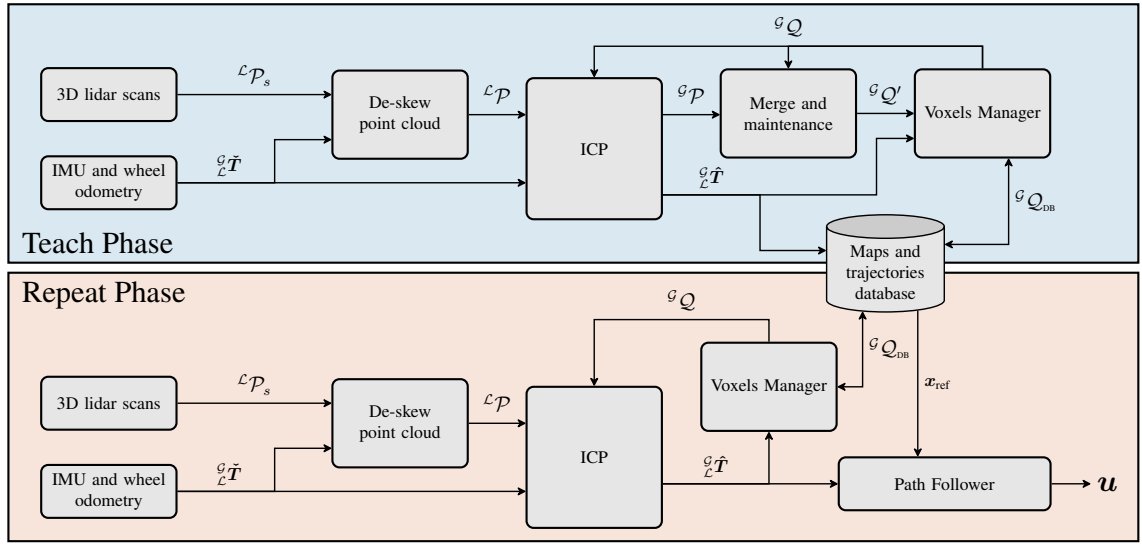

**Figure 6.** Overview of the WILN pipeline. The teach phase (in blue) takes as input the point clouds from the lidar and the odometry. The information is then used by the deskewing algorithm and sent into ICP, which is used to solve the SLAM problem. At the end of the teach phase, the local $^\mathcal{G}\mathcal{Q}$ and nonlocal $^\mathcal{G}\mathcal{Q}_{DB}$ maps are saved in the database. The inputs of the repeat phase (in red) are the same as for the teach phase. The transformation outputted by ICP is finally used by the path follower to compute the commands sent to the vehicle.

$^\mathcal{G}\mathcal{Q}$. This module is also used to compute surface normals and remove dynamic points from the map, yielding the maintained map $^\mathcal{G}\mathcal{Q}'$. Then, the voxel manager module is used to split the map into the latest local map $^\mathcal{G}\mathcal{Q}$ and nonlocal map $^\mathcal{G}\mathcal{Q}_{DB}$. At the end of the teach phase, the entire map and reference trajectory $x_{\text{ref}}$ are saved in the database for later use in the repeat phase. For the repeat phase, the ICP algorithm registers incoming lidar scans identically as in the teach phase. However, only the estimated robot pose $^\mathcal{G}_\mathcal{L}\hat{T}$ is used. The reference trajectory $x_{\text{ref}}$ and current robot pose $^\mathcal{G}_\mathcal{L}\hat{T}$ are then sent to the path follower, which outputs the command vector $u$. Since no point clouds are appended to the map, no map maintenance module is used during the repeat phase. The voxel manager module is used to build the local and nonlocal maps; only the former is taken into account by the ICP algorithm.

## 4. Experimental Setup

In this work, we conducted our deployment within the Montmorency boreal forest, located at a latitude of 47°19′15″N and a longitude of 70°9′0″W The deployment took place during winter, creating ideal conditions to evaluate the impact snowfall and dense vegetation can have on autonomous navigation. The hardware used to perform autonomous navigation is weatherproof and can navigate steep and soft terrain. Section 4.1 describes the winter-resilient hardware used to deploy the WILN system, including the UGV, sensing and computing hardware. Next, details about the various implementation parameters are presented in Section 4.2. Lastly, Section 4.3 lists the characteristics of the Montmorency boreal forest, the weather and conditions to which our system was subjected when navigating.

### 4.1. Hardware Description

Our system was deployed on a modified Clearpath Robotics Warthog UGV, as shown in Figure 7. The Warthog is a SSMR using two drive units located on each side of its chassis. For SSMRs, steering is done by rotating the wheels on each side of the vehicle at different velocities to create a skidding effect, effectively turning the vehicle. Vehicle motion control is achieved through a sub-servo system allowing each side's wheel velocity to controlled through Sevcon Gen4 drives and wheel encoders signal. A kinematic linear mapping between wheel velocities and body velocities allows body-velocity commands to be sent directly to the platform. Instead of wheels, the Warthog is mounted on four CAMSO ATV T4S tracks to maximize mobility, as depicted in Figure 7. The Warthog is also equipped with a differential suspension, increasing traction when navigating steep terrain. A Robosense RS-32 3D lidar is mounted in front of the robot, with no pitch or roll inclination. This lidar has a 200-m detection range and produces about 640 000 points per second. Three Hall effect sensors are added to each motor to provide wheel odometry for the robot. Completing the WILN sensor package, an XSens MTi-10 IMU provides angular velocity and body linear acceleration measurements. Additionally, two Emlid Reach-RS + RTK GNSS receivers were added to the robot chassis and a third was used as a fixed antenna to produce GNSS localization measurements. An Acrosser AIV-Q170V1FL onboard computer is used to record sensor data and perform all of the WILN system computations. All technical specifications for the platform are given in Figure 7.

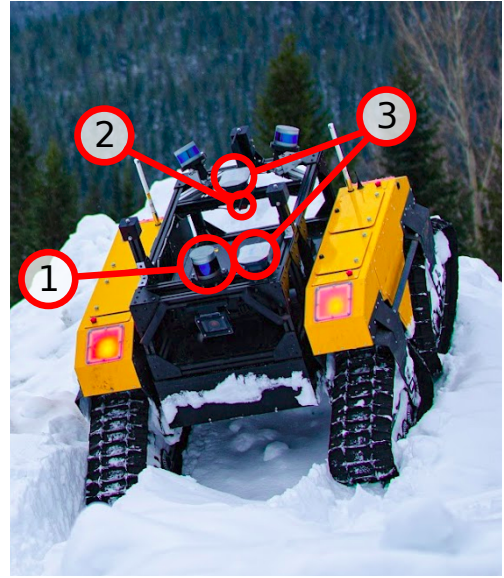

|  | Characteristics | Value |
|---|---|---|
| Physical | Mass | 590 kg |
|  | Footprint | 2.13 m x 1.52 m |
|  | Top speed | 18 km/h |
|  | Steering geometry | Skid-steering |
|  | Locomotion | CAMSO ATV T4S Tracks |
|  | Suspension | Geometric Passive Articulation |
| Sensors | Lidar | Robosense RS-32 (10 Hz) |
|  | IMU | XSens MTi-10 (100 Hz) |
|  | Wheel encoders | 3 × Hall effect sensors (4 Hz) |
|  | GNSS | Emlid Reach-RS+ (5 Hz) |
| Computing | Computer | Acrosser AIV-Q170V1FL |
|  | CPU | i7-6700 TE (3.40 GHz) |
|  | ↳ Number of cores | 4 |
|  | ↳ Number of threads | 8 |
|  | ↳ RAM | 16 Gb |

**Figure 7.** Experimental setup for the WILN system on our Clearpath Robotics Warthog UGV. Left: Detailed specifications. Right: The numbers in red circles correspond to (1) Robosense RS-32 lidar, (2) XSens MTi-10 IMU, and (3) two Emlid Reach-RS+ GNSS antennas. The two GNSS antennas are not used for the localization of the WILN system.

### 4.2. Implementation Parameters

To allow the WILN framework to perform large-scale navigation, we hand-tuned the parameters of each subsystem. All parameters are enumerated in Section 3. Without conducting extensive calibration, we let the system repeat several short paths and manually adjusted parameters until observing stable performance. For this work, parameters were tuned to reach a working state for the WILN system, but no sensitivity analysis was conducted. Our goal was to evaluate the impact of the boreal forest and winter conditions on the system without extensive effort on system calibration. The mapping subsystem had three distinct goals: (i) enabling point cloud registration by computing surface normals, (ii) maintaining the map to remove dynamic points, and (iii) splitting the map into local and nonlocal through the voxel manager. This would prevent localization failures during repeat runs. Path-following overshoot in tight curves and controller oscillations were also minimized by adjusting the parameters. An overview of the tuned parameters is provided in Table 1.

**Table 1.** Parameter values related to each function used in our system. No extensive tuning was conducted, rather we have identified a minimal working system. Functions and parameters are split between the registration, mapping, and path-following subsystems. All functions and parameters are detailed in Section 3.

| | Function | Parameters |
|---|---|---|
| Registration | Bounding box input filter 1 | $b_1 = [-1.5\,\text{m}, 0.5\,\text{m}, -1\,\text{m}, 1\,\text{m}, -1\,\text{m}, 0.5\,\text{m}]$ |
| | Bounding box input filter 2 | $b_2 = [-10\,\text{m}, -1.5\,\text{m}, -2.5\,\text{m}, 2.5\,\text{m}, -1\,\text{m}, 1\,\text{m}]$ |
| | Random sampling input filter | $\eta_s = 0.7$ |
| | Maximum radius input filter | $r = 80\,\text{m}$ |
| | KD tree matcher | $n_m = 7 \quad \varepsilon = 1 \quad d_{\max} = 2.0\,\text{m}$ |
| | Trimmed distance outlier filter | $\eta_d = 0.7$ |
| | Differential transformation checker | $\varepsilon_{\theta_{\min}} = 0.001 \quad \varepsilon_{t_{\min}} = 0.01$ |
| | Counter transformation checker | $i_{\max} = 40$ |
| Mapping | Large scale mapping | $\rho = 0.1\,\text{m} \quad v_s = 20\,\text{m}$ |
| | Surface normal points filter | $n_n = 15$ |
| | Dynamic points filter | $\tau_d = 0.8$ |
| Path-following | Waypoint tolerance | $\tau_w = 1.0\,\text{m}$ |
| | Goal tolerance | $\tau_g = 0.15\,\text{m}$ |
| | Regulator path convergence | $k = 0.4$ |
| | Compensation angular speed command | $K_h = 3.0$ |
| | Max angular velocity | $\omega_m = 1.0\,\text{rad s}^{-1}$ |
| | Goal position factor | $K_g = 0.5$ |
| | Linear nominal speed | $v_{\text{nom}} = 1.5\,\text{m s}^{-1}$ |
| | Linear minimal speed | $v_{\min} = 0.5\,\text{m s}^{-1}$ |
| | Linear maximal speed | $v_{\max} = 1.5\,\text{m s}^{-1}$ |
| | Reference trajectory pose distance | $d_{\text{ref}} = 5\,\text{cm}$ |

### 4.3. Environment

For this work, autonomous navigation was conducted in the Montmorency boreal forest. This environment is ideal to help highlight how dense vegetation and snow precipitation affects the performance of lidar and GNSS-based autonomous navigation, which are the main contributions of this paper. A digital terrain model of the deployment area, with a representation of the path network, is shown in Figure 8. It can be seen that three different paths were defined, all linking two points of interest (POIs). The goal of defining three different paths is to highlight the difference in localization performance between the two path types. Path $A$ links the *Garage* and *Cabin* POIs through a cross-country ski trail and has the total distance of 1.4 km. According to guidelines, we expect this ski trail to have a minimal width of 4.5 m (Ministère des Forêts, de la Faune et des Parcs, 2017). Path $B$ also links the *Garage* and *Cabin* POIs and has a total distance of 1.5 km. Path $B$ is identical to path $A$ for the first third of the path and then diverts to a foot trail, where the width is similar to the one observed on path $A$. We observed that multiple areas of path $B$ have a lower width than the expected 4.5 m since this is a foot trail. To prevent UGV immobilization, the snow on paths $A$ and $B$ had been compacted by a snowmobile operator prior to the experiment. Path $C$ connects the *Garage* and *Gazebo* POIs mostly through a network of wider roads and has a total distance of 0.5 km. Path $C$ corresponds to a forest road, with a minimum width of 9.1 m according to official guidelines of the Montmorency boreal forest (Ministère des Forêts, de la Faune et des Parcs, 2021). It should be noted that path $C$ was conducted as a round-trip path, consisting of a forward pass and a backward pass. This choice was made to increase autonomous navigation data and include data with the UGV driving in both directions. Also, it can be observed in the above mean sea level (AMSL) model in Figure 8 that the path network is located in a valley, surrounded by various mountains.

During the deployment week, the WILN system was subjected to fluctuating weather, including light and heavy snow, hail, and drizzle. We have access to extensive data gathered through a meteorological station located at the Montmorency forest. The area temperature was measured with a temperature probe of model 107 made by Campbell Scientific, which was covered by an antiradiation screen 2 m away from the ground. The precipitation type was measured with a Parsivel$^2$ disdrometer made by OTT. An overview of the meteorological conditions is illustrated in Figure 9. Outside temperatures ranged from $-15.5\,°C$ to $13.4\,°C$. The temperature fluctuated over and under the water freezing point (i.e., $0\,°C$), meaning that snow in the environment melted on some occasions and froze on others, effectively accumulating on the ground. Teach runs (i.e., TA, TB, and TC) were conducted when there was no precipitation. Runs R1 through R7 were conducted in rainy weather, while runs R8 through R12 were conducted in snowfall. The standard for weather codes shown in Figure 9 is code table 4680 for automatic weather stations (World Meteorological Organization, 2019). During runs R13 and R14, the system was not subject to any precipitation. While not shown in this figure, we performed autonomous repeats during both daytime and nighttime, with no impact on system performance. More details on each run are given in Section 5.

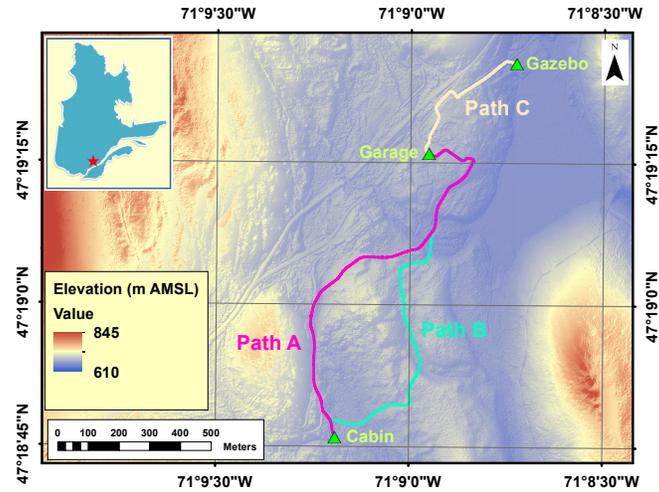

**Figure 8.** A digital terrain model of the Montmorency forest, where the WILN system was deployed, and the forest location in Québec, Canada (top left). We see the three different paths, both paths *A* (1.4 km) and *B* (1.5 km) going from the *Garage* points of interest (POI) to the *Cabin* POI, while path *C* (0.5 km) connects the *Garage* POI to the *Gazebo* POI. Global latitudinal and longitudinal coordinates are given in the margin. The coloring is defined by Above Mean Sea Level (AMSL) elevation. Image credit: Remote and Wood Sensing Laboratory, University of New Brunswick.

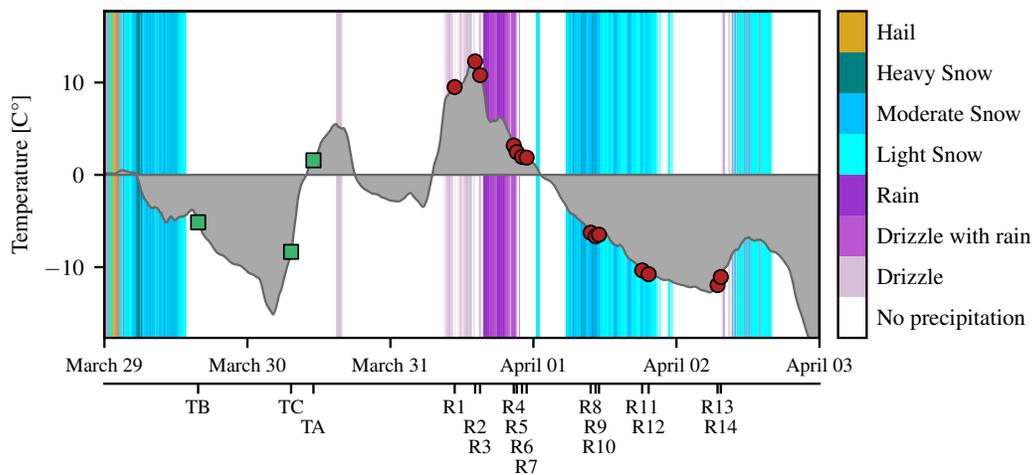

**Figure 9.** Temperature and weather during the deployment. Teach runs are depicted as green squares whereas repeat runs are depicted as red circles. Teach runs have been performed in relatively clear weather, while repeat runs suffered from rain and different levels of snowfall.

## 5. Results

The main goal of this work is to evaluate the impact of the boreal forest and winter weather on autonomous navigation performance. First, we document all three distinct reference maps and trajectories that were built through the teach phase. Then, we report on all repeat runs that were conducted over the deployment week. As a prior to further analysis, general information on all runs conducted during the deployment is presented. A general performance report for the system is shown in Section 5.1. Afterwards, we highlight the effects of the boreal forest biome, more specifically of the dense vegetation, on GNSS-based and ICP-based localization in Section 5.2. Then, Section 5.3 describes how snow precipitation and accumulation induces significant change in the environment landscape, resulting in ICP localization failures. Finally, Section 5.4 explains the impact of the environment on the OE controller path-following performance.

Teach runs were performed to record the reference maps $^{\mathcal{G}}\mathcal{Q}_{\mathrm{DB}}$ and the trajectories $\boldsymbol{x}_{\mathrm{ref}}$ for all three paths. During these runs, we manually drove the robot, while the map and trajectory database were built online. The resulting maps for each path are illustrated in Figure 10 with the coordinates defined in the map frame $\mathcal{G}$ specific to each path. In this visualization, the $z$ component of the estimated surface normal vector is used to color each point. Thus, dark blue points represent walls, as can be seen on path $C$, green points typically represent vegetation, and yellow points represent ground surfaces. Paths $A$ and $B$ are mostly located within dense vegetation. On the other hand, path $C$ is located on a much more open forest road. All reference paths in Figure 10 have been divided into areas to facilitate upcoming discussion. Table 2 presents relevant details about each

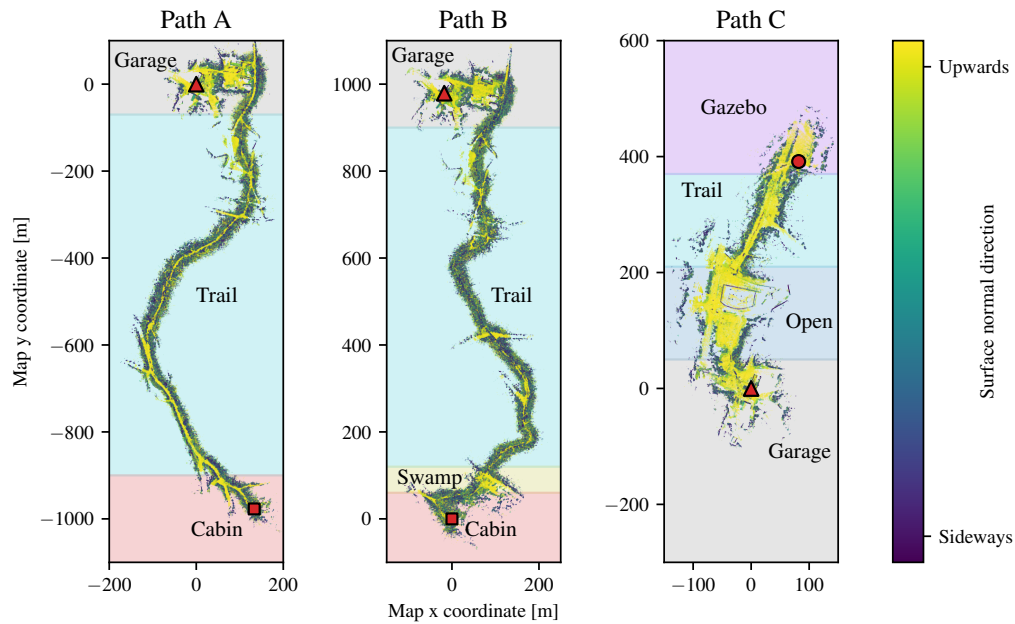

**Figure 10.** All reference maps recorded during the three teach phases for this work. The colormap represents the $z$ component of the surface normal for all points in the map. Yellow points are the ones with the surface normal pointing upwards. Green and dark blue points have a surface normal pointing sideways. The red markers represent the reference trajectory POIs. All maps have been rotated in order to allow visualization in the same figure. The path areas identified here are used throughout the remainder of this section. The red markers represent the POIs associated to each path. The origin of path $B$ is located at the *Cabin* POI.

1644 · Baril et al.

**Table 2.** Overview of all runs conducted in this work. All times are defined in the local Eastern Standard Time. A' or B' means the run started at the *Cabin* POI. The column $\Delta t$ defines the elapsed time since the teach run of the associated path. The column $\#\mathcal{P}$ shows the number of scans for each run. The illumination measured at the start of each run is given. Teach runs are not accounted for in the total autonomous distance traveled.

| ID | Path | Start time (2021) | Duration (hh:mm) | $\Delta t$ (hh:mm) | $\#\mathcal{P}$ | Distance (km) | Illumination (W m$^{-2}$) | Interventions |
|---|---|---|---|---|---|---|---|---|
| TB  | B' | March 29th 15:46 | 00:27 | 00:00 | 15 715 | –   | 283.35 | – |
| R2  | B  | March 31st 14:12 | 00:27 | 46:26 | 16 183 | 1.5 | 295.10 | 0 |
| R5  | B' | March 31st 21:14 | 00:27 | 53:28 | 16 056 | 1.5 | 0.00   | 0 |
| R6  | B  | March 31st 22:05 | 00:27 | 54:19 | 16 163 | 1.5 | 0.00   | 0 |
| R9  | B' | April 1st 10:27  | 00:27 | 66:41 | 16 093 | 1.5 | 123.90 | 2 |
| R11 | B  | April 1st 18:18  | 00:32 | 74:32 | 19 443 | 1.5 | 0.75   | 1 |
| R14 | B' | April 2nd 07:28  | 00:27 | 87:42 | 16 073 | 1.5 | 188.42 | 0 |
| TC  | C  | March 30th 07:29 | 00:21 | 00:00 | 12 762 | –   | 102.58 | – |
| R1  | C  | March 31st 10:46 | 00:23 | 27:17 | 13 758 | 1.0 | 284.80 | 0 |
| R10 | C  | April 1st 11:03  | 00:20 | 51:34 | 12 162 | 1.0 | 159.23 | 0 |
| TA  | A  | March 30th 11:06 | 00:25 | 00:00 | 15 142 | –   | 586.38 | – |
| R3  | A' | March 31st 15:04 | 00:26 | 27:58 | 15 535 | 1.4 | 101.55 | 0 |
| R4  | A  | March 31st 20:44 | 00:26 | 33:38 | 15 641 | 1.4 | 0.00   | 0 |
| **R7** | A' | March 31st 22:49 | 00:15 | 35:43 | 8862 | **0.8** | 0.00 | 1 |
| R8  | A  | April 1st 09:25  | 00:28 | 46:19 | 15 623 | 1.4 | 84.68  | 0 |
| R12 | A' | April 1st 19:19  | 00:29 | 56:13 | 17 218 | 1.4 | 0.00   | 0 |
| R13 | A  | April 2nd 06:55  | 00:26 | 67:49 | 15 583 | 1.4 | 99.63  | 0 |
| Total | 14 runs | – | 07:13 | – | 258 012 | 18.8 | – | 4 |

of the teach and repeat runs conducted during the deployment. A total of 14 repeat runs were completed, summing up to 18.8 km and 7 h 13 min of autonomous repeating over 5 days. The last repeat run was started over 87 h after its respective teach run on path B. A battery power outage prevented us from completing run **R7**; it was therefore interrupted. The system also suffered from three localization failures in runs R9 and R11. For each run, the sun radiation measured by a CNR4 radiometer equipped with a CNF4 ventilation unit made by Kipp & Zonen located within the forest is given.

### 5.1. General Performance Report

This section aims to give a general overview of our system's performance and show that we achieved enough autonomy to generate observations on navigation in boreal forests. To characterize the reliability of our system, we computed the cross-track error, which is the distance between the robot frame $\mathcal{R}$'s origin and its orthogonal projection on the path, as defined by Mondoloni et al. (2005). The cross-track error for every run is shown in Figure 11. It can be seen that this error generally stays below 1.0 m for all runs, with some peaks corresponding to the various turns in each reference trajectory. This error is below the cross-country trail half-width of 2.25 m, as specified in Section 4.3. Note that UGV initialization was done using visual markers in the *Garage* area, but not in the *Cabin* area. The resulting high initialization error explains the high cross-track error at the end of paths A and B. Additionally, it can be seen in the last column of Table 2 that a total of four manual interventions were done through the deployment. The three interventions done through runs R9 and R11 were due to localization failure. The localization failure on run R11 occurred in the *Trail* area of path B. At this point, a large drift in localization caused the robot to divert from the trail and hit a tree, leading to a cross-track error peak, as can be seen in Figure 11. Two more localization failures occurred in the *Cabin* area of path B during run R9. For these failures, it can be seen in Figure 9 that run R9 occurred during a snowstorm, leading to significant snow accumulation on the ground. An in-depth analysis of this event is presented in Section 5.3. The intervention done in run R7 was due to battery power outage. Excluding the intervention done due to battery power outage, there has been an average of one intervention per 4.7 km traveled during this deployment. Thus, we conclude that the WILN system showed a satisfactory degree of autonomy to generate observations on autonomous navigation in boreal forests.

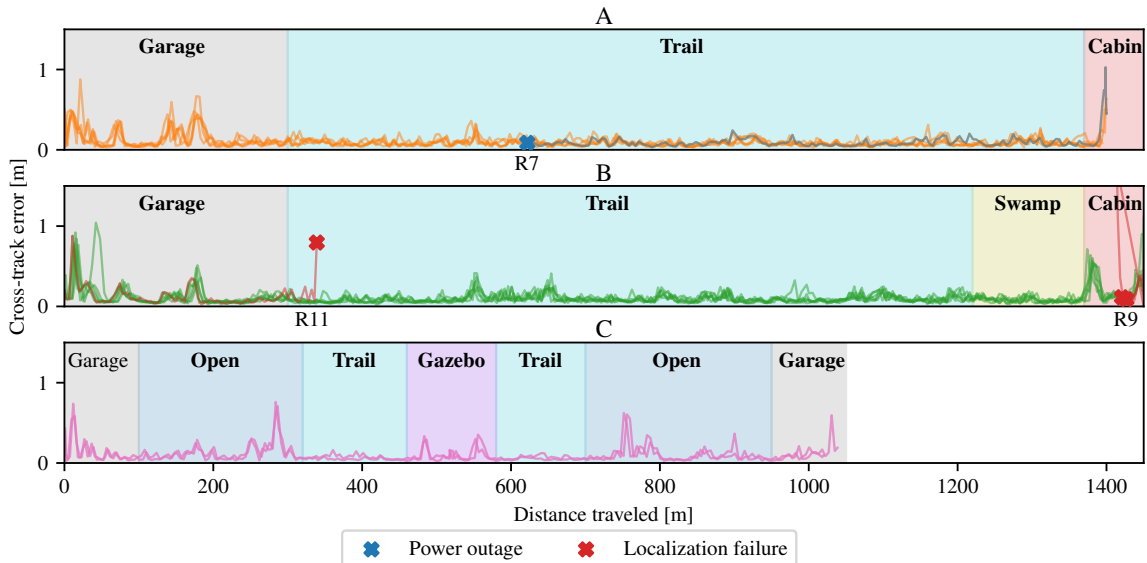

**Figure 11.** Cross-track error with respect to distance traveled for all runs. The results from paths *A*, *B*, and *C* are plotted in orange, green, and pink, respectively. The three localization failures from runs R9 and R11 are depicted as red crosses. Run R7, which ended prematurely due to a battery outage, is depicted as a blue cross. The shaded areas for each path correspond to those shown in Figure 10.

### 5.2. Impact of the Boreal Forest Biome

The first main goal of this work is to evaluate the impact of dense vegetation inherent to boreal forests on both GNSS and lidar-based localization for autonomous navigation. During all repeat runs, the GNSS signal statistics were logged in both receivers mounted on the Warthog UGV. Additionally, measurements were taken from a fixed reference receiver during each run to enable RTK-GNSS positioning. The data were post-processed offline using RTKLIB[5] (Takasu and Yasuda, 2009) to produce the most accurate measurements possible. To estimate the GNSS localization error, we compared the manually measured distance between the GNSS receivers mounted on the UGV with the distance indicated by the GNSS localization, similarly as in Vaidis et al. (2021). This metric does not take into account localization biases to which both receivers might be subjected; thus we consider the error metric an optimistic one. The results of the GNSS error with respect to the mean number of satellites between the two mobile receivers are presented in Figure 12. Our dataset does not contain a sufficient number of data points with less than seven satellites; we therefore exclude them from our analysis. This shows that for reliable GNSS localization on forest trails, the minimum number of observed satellites, $n_{cs}$, is 18. If a lower number of satellites is observed, the system is at risk of collision with the vegetation at the edges of the path. Since the off-road paths are wider, they allow slightly lower $n_{cs}$ of 16.5 for safe navigation.

Based on the critical number of satellites determined in Figure 12, we have conducted an analysis of all reference trajectories to determine the areas considered to be GNSS denied. Figure 13 shows the mean number of observed satellites in a georeferenced satellite image. On paths *A* and *B*, the mean number of satellites drops below the critical number $n_c$ at several locations, mostly in areas located within dense vegetation. The mean number of observed satellites increases over $n_c$ in the open areas of the trail network. Path $C$ is the only path that could be repeated reliably using GNSS localization. This is due to the fact that its environment is considerably more open compared to $A$ and $B$, as shown in Figure 10. Additionally, it is necessary to stress that to achieve the presented accuracy in real time, a stable data link between the mobile GNSS receivers and the reference receiver must be maintained. In practice, this data link is usually implemented by a radio link or by a mobile internet connection, which can be complex to set up in remote locations. In our case, the reference receiver radio signal could not cover the entirety of the *A* and *B* paths. Yet, we confirm that the reference receiver positioned on an elevated spot which is not occluded by trees (i.e., A1) provides better range compared to the locations A2 and A3 in the forest.

---
[5] http://www.rtklib.com/

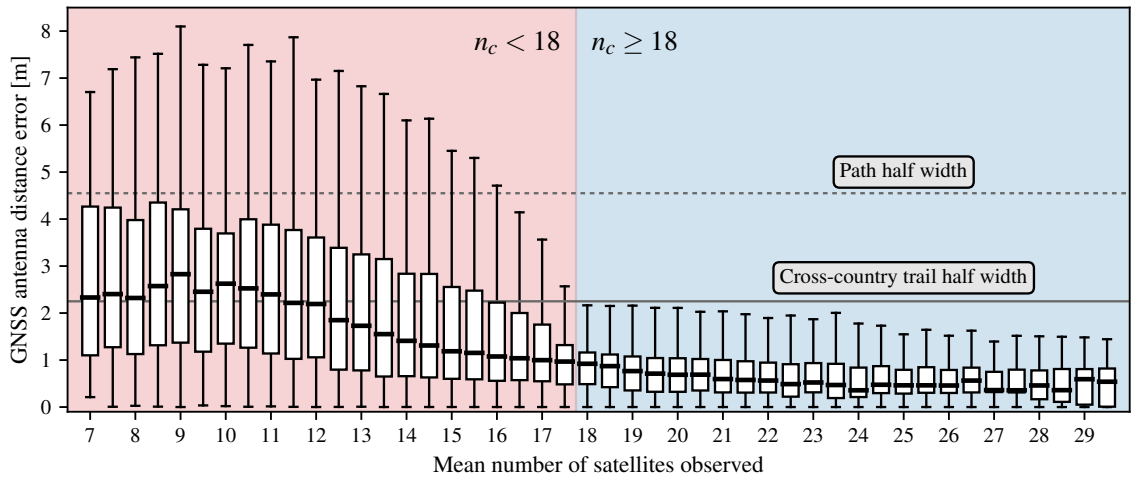

**Figure 12.** GNSS error with respect to the mean number of observed satellites between the two receivers mounted on the UGV over all runs. The thick black lines represent medians, the boxes represent the first and third quartiles, and the whiskers represent the 10th and 90th percentiles. The expected forest road and forest trail half-widths are 4.55 m and 2.25 m, respectively. The shaded blue area denotes the acceptable number of observed satellites for reliable navigation in wood trails. The area in red is unacceptable for forest trail navigation, based on the critical threshold $n_c$.

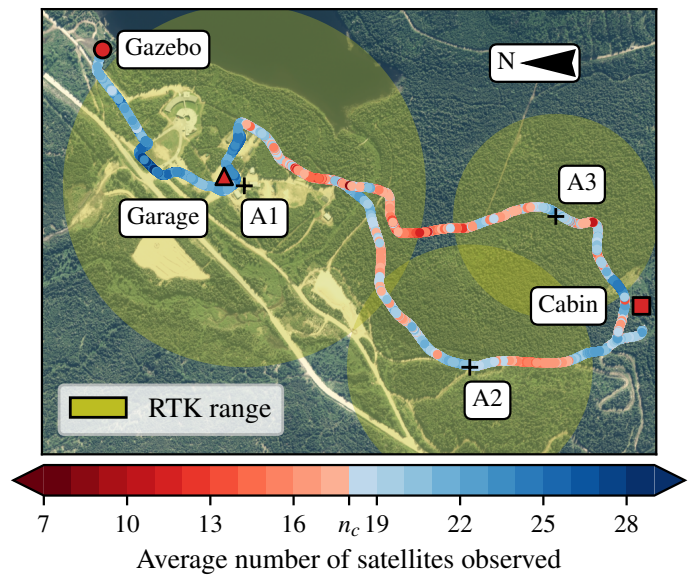

**Figure 13.** GNSS satellites coverage along the reference trajectories. The blue palette is used for the sections where the GNSS accuracy would be sufficient for the width of the trail. The red sections would not allow pure GNSS navigation. For RTK-GNSS localization, the distinct positions of the reference receiver are the A1, A2, and A3 points. The radio signal range of the reference receivers is shown in olive. Image credit: Forêt ouverte.

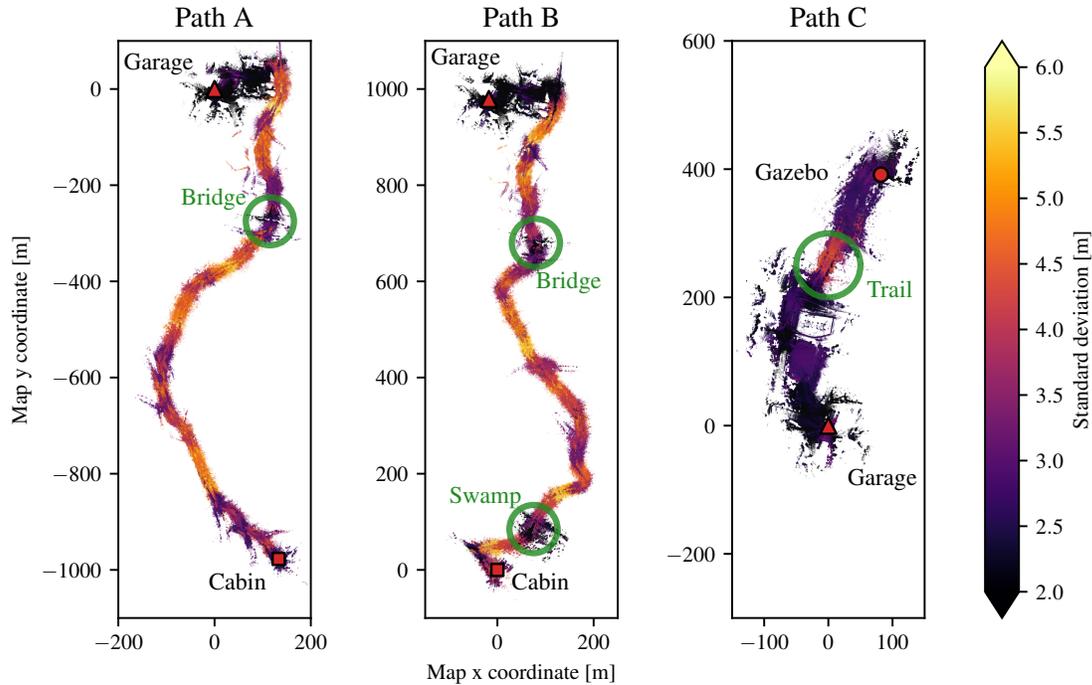

**Figure 14.** Top view of a set of 500 lidar scans $^{\mathcal{G}}\mathcal{P}$ from runs R4, R5, and R1, conducted on paths *A*, *B*, and *C*, respectively. All of the scans are distributed evenly time-wise in their respective runs. The standard deviation of the point-to-plane error e after applying perturbations is used to color each scan. A higher standard deviation is linked to a higher localization uncertainty. Distinctive areas are highlighted by the green circles. The starting and ending POIs for each trajectory are denoted by the red markers.

Furthermore, the ICP algorithm provided localization accurate enough to allow the robot to complete each repeat run without colliding with obstacles. Yet, we observed that the corridor-like nature of the boreal forest trails leads to a high localization uncertainty. To investigate this phenomenon, we extracted 500 lidar scans $\mathcal{P}$ from runs R4, R5, and R1, uniformly distributed time-wise, all of which are shown in Figure 14. These runs are the first repeat runs conducted in the same direction for each reference path. For each lidar scan $\mathcal{P}$, we expressed the map $\mathcal{Q}$ in the corresponding lidar frame $\mathcal{L}$. The initially well-registered scans were perturbed along the robot's longitudinal axis by translations from $-6\,\text{m}$ to $6\,\text{m}$ with increments of $0.05\,\text{m}$. The point-to-plane error e, defined in Equation 2, was then evaluated for each perturbed scan with respect to the reference $\mathcal{Q}$. It should be noted that the ICP algorithm was not executed to convergence for the analysis. The unregistered error was computed for each perturbed point cloud. The standard deviation of the error was used to color each scan.

The areas located within the cross-country skiing trails are related to a considerably higher standard deviation. Indeed, the ICP error function, presented in Equation 2, is much flatter in such areas. A pattern emerges: the standard deviation increases when the robot traverses long and narrow forest corridors, and then decreases when driving through trail intersections. Moreover, the standard deviation in the *Bridge* area, located in paths *A* and *B*, decreases notably compared to its surroundings. The river breaks the line of trees in this area and provides additional geometrical constraints, effectively improving the localization accuracy. A similar effect can be observed in the *Swamp* area in path *B*. A considerable amount of vegetation is perpendicular to the longitudinal direction of the robot, better constraining the registration error function. Analogously, the results from path *C* demonstrate that the standard deviation is much lower in the wide forest paths. The only place where the localization uncertainty increases again is the narrow trail area marked by the green circle.

### 5.3. Impact of Snow Precipitation and Accumulation

Another important goal of the deployment is to study the impact of winter weather on autonomous navigation. Paton et al. (2017) already mentioned illumination variation, low feature contrast, and a changing environment as issues for vision-based navigation. Thus, in this work, we focus on the impact of snow precipitation on lidar-based localization. While the system was deployed under various weather conditions, as shown in Figure 9, we did not observe any correlation between the precipitation and the immediate accuracy of localization. However, the snow accumulation led to significant changes in the environment, which affected the ability of our system to localize between teach and repeat runs. More specifically, the WILN system failed to initialize its localization in the reference map during two attempts to start run R9. The third attempt was successful, and the system repeated the entire path without fail subsequently. As documented by Figure 9, run R9 was conducted right after moderate snowfall, and there was thus a significant amount of snow accumulation in the environment.

To analyze the impact of snow accumulation on the reliability of the lidar-based localization, we extracted 500 lidar scans $^{\mathcal{G}}\mathcal{P}$ from runs R5, R9, and R14. These scans were already registered with the map and are distributed evenly with respect to time. All three runs were conducted on path *B* from the *Cabin* POI to the *Garage* POI. Then, we evaluated the overlap percentage for each $^{\mathcal{G}}\mathcal{P}$ scan, presented in Figure 15. A scan point was considered as overlapping the map if it was located closer than 0.5 m to any $\mathcal{Q}$ point.

It is apparent that the scan overlap in run R9 is considerably lower than the scan overlap in R5 and R14, except for the *Garage* area, where R14 has the lowest scan overlap. For the *Cabin*, *Swamp*, and *Trails* areas, the lower overlap ratio can be attributed to snow that accumulated in the environment, which was significantly higher in R9 due to snowfall. Specifically for the *Cabin* area, the large drop in the scan overlap explains the multiple failed initialization attempts mentioned earlier. Furthermore, looking at the *Garage* area, it can be seen that scan overlap decreases as time progresses. Indeed, intermittent snow plowing operations were conducted in this area. Growing snowbanks at the edge of the road led to dynamic changes in the area landscape and consequently to a progressive decrease in lidar scan overlap with the reference map, as shown in Figure 15.

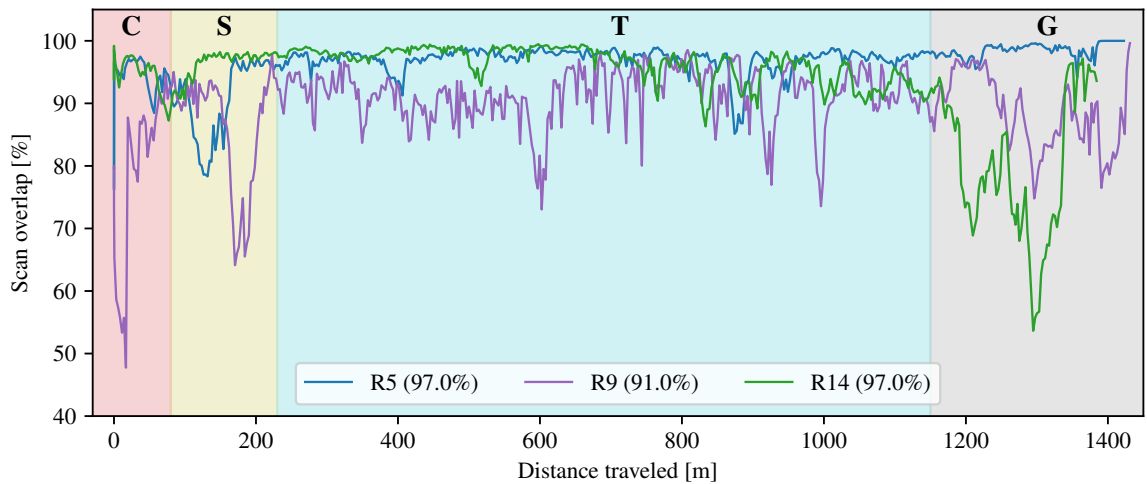

**Figure 15.** Overlapping percentage of lidar scans $^{\mathcal{G}}\mathcal{P}$ on the reference map $^{\mathcal{G}}\mathcal{Q}$. The results from runs R5, R9, and R14 are plotted in blue, purple, and green, respectively. The median scan coverage for each run is written between parentheses in the legend. The shaded **C** area represents the *Cabin* area, **S** represents the *Swamp* area, **T** represents the *Trails* area, and **G** represents the *Garage* area.

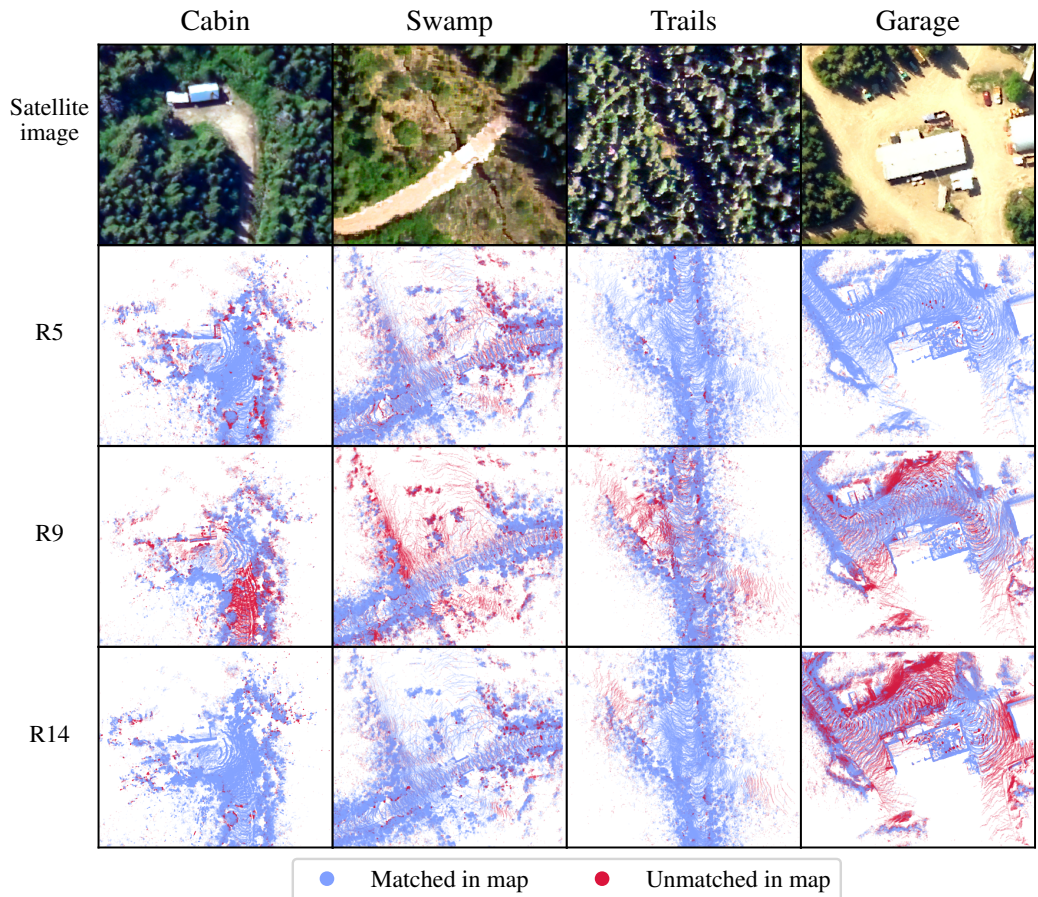

**Figure 16.** A detail of various parts of path *B*. The satellite images in the top row illustrate the locations. In the registered lidar scans from runs R5, R9, and R14, the points are colored blue if they have a counterpart in the reference map. Contrarily, red points miss their map counterpart and mark a new feature or a change in the environment.

As specific examples, we extracted four problematic areas related to snow accumulation, all of which are shown in Figure 16. A red color mask was applied to the lidar scans $\mathcal{P}$ to highlight the nonoverlapping points. In the *Cabin* area during run R9, it is apparent that snow accumulation led to a variation in the terrain steepness, which affected the ability of the ICP algorithm to initialize its localization within the reference map. The difference in the scan overlap between run R9 and other runs is also noticeable in the *Swamp* and *Trail* areas, although to a lesser extent. Following what is shown in Figure 15, the overlap percentage in the *Garage* area was gradually decreasing as time passed. This phenomenon was related to the aforementioned snow plowing operation taking place in that area, effectively changing the snow landscape. During run R14, there was also a large truck parked near the *Garage*, which contributed to the unmatched points. It should be noted that while scan overlap reaches low values in the *Garage* area, the WILN system did not suffer from localization failure during any run in this area. The location contains multiple buildings, resulting in a higher number of geometrical constraints, allowing the ICP algorithm to localize the robot despite variations in landscape. We assume that the scan overlap would continue to decrease as more time is elapsed between a repeat run and its respective teach run.

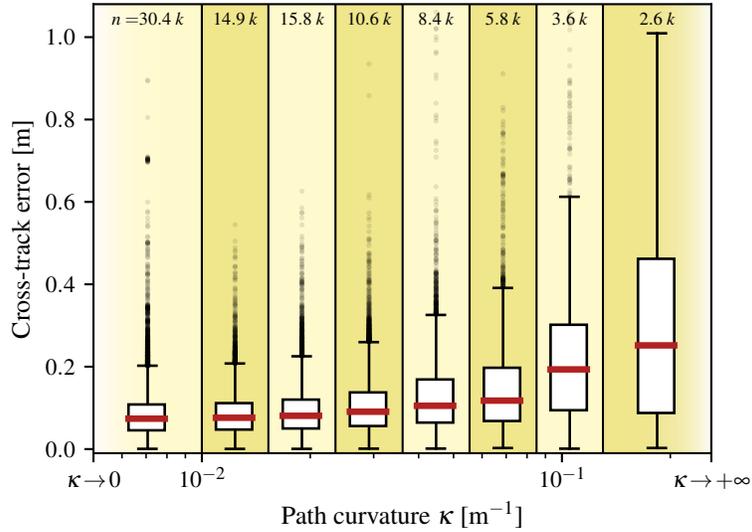

**Figure 17.** Cross-track error as a function of path curvature for all repeat runs. All errors were split into bins representing different path curvatures. The thick red lines represent medians, the boxes represent the first and third quartiles, respectively, and the whiskers represent the 10th and 90th percentiles. The path curvatures on abscissa are defined in a logarithmic scale. All data related to path curvatures smaller than $0.01\,\text{m}^{-1}$ are grouped in the first bin. Similarly, all data related to path curvatures greater than or equal to $0.13\,\text{m}^{-1}$ are grouped in the last bin.

### 5.4. Path-Following Performance

In this section, we have concatenated the cross-track error for all runs; the results can be seen in Figure 17. The median cross-track error across all $A$, $B$, and $C$ path runs is 0.083, 0.088, and 0.075 m, respectively. The maximum observed cross-track errors across all runs is 1.19 m for path $A$, 1.17 m for path $B$, and 0.92 m for path $C$. These results suggest that the path-following performance slightly decreases when navigating on a forest trail (i.e., paths $A$ and $B$), compared to the forest road (i.e., path $C$). When comparing the results, we did not observe a correlation between weather conditions and cross-track error.

Nonetheless, we did observe significant correlation between path curvature and cross-track error. To analyze this phenomenon, we estimated the path curvature $\kappa$ of each reference trajectory point. To remove localization noise from the curvature computation, we took ten nearest reference trajectory points into account. We then subsampled all repeat trajectories to keep points at a distance of 0.1 m from each other. For each point of the subsampled trajectories, we found the closest reference trajectory point, computed the cross-track error $\epsilon_{CT}$, and associated it with the reference point's curvature $\kappa$. This computation allows us to correlate path curvature with cross-track error, as shown in Figure 17.

We detect that cross-track error increases proportionally to the path curvature. For low curves (i.e., the first six bins), the 90th percentile of the cross-track errors remains below 0.6 m. For straight lines (i.e., the first bin), we attribute the excessive errors over 0.6 m to the initialization error in the beginning of the repeat runs. When navigating in moderate turns (i.e., the last two bins), the 90th percentile of cross-track errors reaches 1.0 m. Due to limitations of the OE controller, the system did not navigate in tight curves (i.e., turning on the spot). The high motion prediction inaccuracy of the kinematic differential-drive model when turning at high angular velocities is the cause for this high increase in cross-track error, as highlighted by Baril et al. (2020). While this error is below the cross-country trail half-width of 2.25 m, which we consider the limit for safe navigation in forest trails, it remains high. It should be noted that the repeat runs were conducted at a conservative nominal velocity $v_{\text{nom}}$ of 1.5 m/s. Therefore, there is a significant risk of hitting vegetation for autonomous systems relying on kinematic controllers when navigating in a boreal forest at higher nominal velocities.

## 6. Challenges and Lessons Learned

In this section, we use the results presented in Section 5 to elaborate how autonomous navigation algorithms should be improved to enable multiyear autonomy. We first explain the challenges related to navigating tight forest trails with dense vegetation in Section 6.1. We also discuss how localization algorithms should adapt to dynamic environments to be deployed all year long in northern environments in Section 6.2. Moreover, we discuss the limitations of kinematic and time-invariant path-following controllers and the risk associated with robot immobilization when navigating in deep snow or through multiple seasons in Section 6.3. Finally, Section 6.5 presents the lessons learned through this field report.

### 6.1. The *Forest Corridor* Effect

We have highlighted the impact of the dense vegetation on GNSS signal reception in Section 5.2. We observed that GNSS-based localization enables autonomous navigation on forest roads (i.e., path $C$), but the signal is not reliable when navigating boreal forest trails (i.e., paths $A$ and $B$). Indeed, the GNSS error can reach upwards to 8 m in forest trails, as can be seen in Figure 12. This effect is due to the signal being absorbed by the dense vegetation in forest trails, an example of which is shown in Figure 1b. Additionally, fusing GNSS measurements with lidar-based localization was investigated by Babin et al. (2019). The authors mentioned that a major issue for real-time RTK-GNSS is that the range of the reference receiver is severely affected by the dense vegetation. In this field report, we have observed the same radio signal range issue, as shown in Figure 13. Enabling real-time RTK-GNSS within boreal forest trails would thus require to set multiple reference antennas throughout the environment. Lastly, it is known that GNSS signal can be jammed, making systems relying on this signal easier to disrupt (Ren et al., 2021). Thus, completely autonomous navigation in a boreal forest requires a localization framework that is resilient to GNSS-denied conditions.

Moreover, we showed in Section 5.2 that featureless corridors are challenging for lidar-based localization algorithms. As can be observed in Figure 14, the uniform nature of boreal forest vegetation leads to low geometrical constraints in the robot's longitudinal direction. Low geometrical constraints lead to high registration uncertainty for the ICP algorithm, as highlighted by Gelfand et al. (2003). This lack of features makes the ICP registration cost function flat in the corresponding direction and sensitive to noise in the lidar measurements. We observe a similar phenomenon in the forest trails surrounded by dense vegetation. Since the lidar scans are not dense enough to distinguish between single branches, the trees resemble large blobs of randomly distributed points in the scans and ultimately in the map. Therefore, point cloud degradation is not limited to the large open areas that were mentioned by Ren et al. (2021). In particular, we have observed that point cloud degradation led to the ICP localization to jumps forward of up to 0.5 m. Such jumps in ICP localization lead to instability and eventually to system failure. Also, they disrupt the function of motion controllers, leading to jerky motion. A system aware of this phenomenon could potentially reduce its velocity in the problematic areas to prevent system failure. A more accurate and adaptive motion prediction model, similar to the one discussed in Section 6.3, could also allow prior error to be reduced and improve localization resiliency in forest corridors. We also argue that future work should investigate multimodal localization approaches. System reliability could be greatly improved by relying on a heterogeneous sensor suite.

### 6.2. Impact of Snow Accumulation

Our initial hypothesis was that during snowstorms, the precipitation would affect the localization accuracy of the ICP algorithm. Yet, as highlighted in Section 5.3, we found no correlation between precipitation intensity and localization accuracy. One should note that while our system was subject to moderate snowfall, no similar work documents the impact of whiteout conditions, which would be caused by heavy snowfall and/or blizzard. We argue that a metric to quantify noise caused by precipitation will be useful to compare various field reports and datasets in the literature. Such a metric would allow to compare various localization approaches under extreme conditions. Rather, we discovered that it was the snow accumulation that severely affected the lidar-based localization. On certain surface types, such as the ground or rooftops, the snow keeps accumulating during the snowstorms. Yet on other surfaces (e.g., trees or running water), only a limited amount accumulates before it falls off or melts. Our data show that lidar-based localization in areas located within deep woods suffer from changes in terrain topology, while areas located near buildings are resilient to this phenomenon. Various examples of the structural change in the environment can be observed in Figure 16. Focusing on the *Cabin* area shown in the first column, it is apparent that terrain steepness changes significantly when comparing run R9 with other similar runs. MacTavish et al. (2018) have mentioned that multiexperience vision-based localization suffered from fast scene appearance change, such as what would be caused by a snowstorm. As mentioned in Section 3.4, the reference map is not updated during the repeat phase for the WILN system. Thus, WILN is also subject to localization failure when attempting to localize in deep woods after snowfall; an example of the result is shown in Figure 18. In this case, the significant structural change in the environment leads to erroneous UGV localization and eventually to crashing with vegetation. However, future work should enable dynamic map maintenance using either multiple maps, similarly to what is proposed by Zhang and Singh (2018) and Maddern et al. (2015) or dynamic maintenance of a single map, similarly to Pomerleau et al. (2014). We argue that using a metric similar to the one presented in Section 5.3 would allow to identify significant structural change in the environment, thus requiring map maintenance. Nevertheless, autonomous systems should adapt their reference maps when observing a mismatch between the current sensor measurements and the reference data to ensure a complete, year-long autonomy.

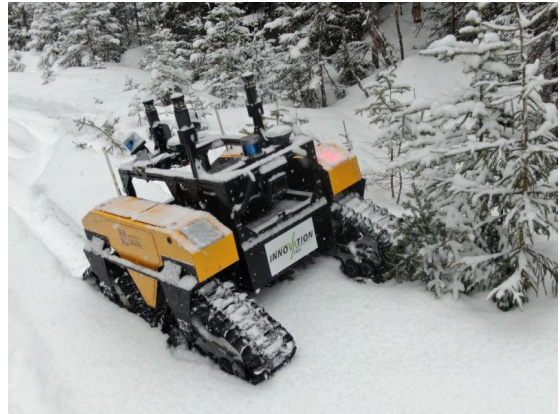

**Figure 18.** The result of a failed localization initialization, during run R9. The high initial error in robot pose resulted in the Warthog diverging from the reference path and heading towards vegetation. The robot reached a 2.25 m deviation before being manually stopped and driven back to the initialization position.

### 6.3. Path Following in Snow-Covered Terrain

In Section 5.4, we described the performance of the OE controller when used on the Warthog SSMR. We showed that when navigating at the target velocity $v_{\text{nom}}$ of 1.5 m/s, this controller is able to repeat kilometer-scale paths, with an error that remains below the half-width of the forest trails. This result is interesting considering that the OE requires little knowledge of the UGV properties and has low computation time. We have observed that path curvature increases cross-track error,

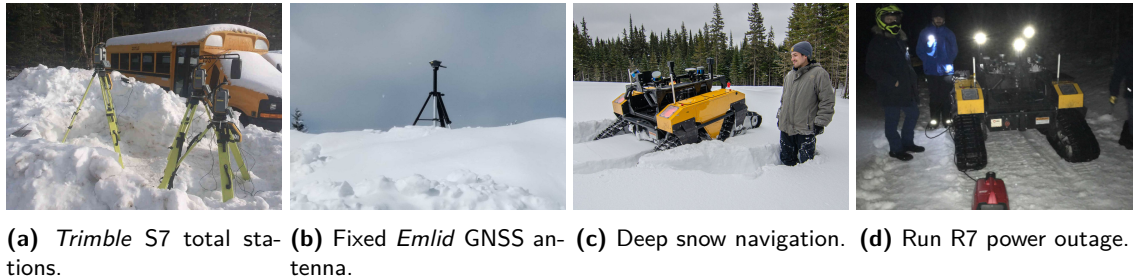

**(a)** *Trimble* S7 total stations.  **(b)** Fixed *Emlid* GNSS antenna.  **(c)** Deep snow navigation.  **(d)** Run R7 power outage.

**Figure 19.** Various lessons learned during this deployment. (a) Total stations that were intended to measure the ground truth localization of the UGV in the *Garage* area. Despite considerable effort to prevent sinkage, those devices shifted significantly, resulting in unusable data. (b) The fixed GNSS antenna to enable RTK-GNSS. This device was placed in open areas to ensure good GNSS signal reception. (c) The Warthog UGV navigating in deep snow. A $1.8\,\mathrm{m}$ tall human operator stands next to the robot for scale. The snow level is estimated at $0.7\,\mathrm{m}$ in this area. The bottom of the robot chassis is resting directly on snow, affecting traction and eventually leading to UGV immobilization. (d) Warthog battery failure during run R7. A generator (in red) was used to recharge the robot batteries and later drive it manually back to the *Garage* POI.

as depicted in Figure 17. This suggests that SSMR motion is difficult to predict when rotating at higher velocities. Thus, we can assume that navigation at higher nominal velocities would potentially lead to the robot crashing into the vegetation. As mentioned in Section 3.5, the original controller implementation presented in Huskić et al. (2017) as well as predictive controllers, such as the one presented in Ostafew et al. (2016), include a parameter reducing vehicle longitudinal velocity when crossing high path curvature. We initially planned to add a similar feature to WILN; however, localization noise in the reference trajectory led to jerking motion in the repeat phase. Reference trajectory smoothing should be investigated in the future to enable speed reduction relative to reference path curvature. Additionally, the OE controller does not react to UGV immobilization. This poses a high risk in off-road navigation, as Figure 19c and Figure 21a demonstrate. A controller based on a more accurate dynamic model that adapts to the various conditions would enable path repeating at higher speeds without failing. A model that can adapt to the differences in UGV behavior across the entire reference trajectory would also improve the path-following resiliency to WILN navigation in steep and soft terrain. Furthermore, WILN is not a completely autonomous system as it does not include obstacle avoidance and planning, as the one described in Krüsi et al. (2015). As future work, it would be interesting to add precipitation-resilient obstacle avoidance. We also argue that obstacle avoidance should include soft ground that prevents the UGV from reaching the next part of the reference path.

### 6.4. Effect of High-Speed Navigation on Localization

During our preliminary tests, we noticed that the WILN system was not robust to quick motions of the robot. When the robot rotated too quickly, the localization would fail because of motion distortions in lidar scans. To solve this issue, we added a point cloud deskewing step, as described in Section 3.1, to correct motion distortions. This solution was sufficient in our case because the robot was driven at the relatively low speed of $1.5\,\mathrm{m\,s^{-1}}$. However, at higher speeds, motion distortion in lidar scans is more important and odometry accuracy is decreased because of slipping, skidding, and potential robot immobilization. To solve this issue, we introduced an algorithm that computes the uncertainty of each point of a lidar scan to know which points can be trusted (Deschênes et al., 2021). Points which are likely to be more affected by motion distortion have a higher uncertainty than others, and thus are given less influence during the registration process. Also, points which are too uncertain can be completely removed from the map to make sure that it stays crisp. This uncertainty-based algorithm, which can be used in combination with deskewing algorithms, would help to preserve localization and mapping precision when navigating at high speeds with the WILN system.

### 6.5. Lessons Learned from Field Deployments

Through the multiple deployments conducted in the completion of this field report, we have learned various lessons that generally apply to deploying mobile robots in northern environments. We have observed multiple issues related to robot navigation in deep snow, to localization uncertainty in forest corridor, and to the winter weather affecting robot autonomy. We argue that autonomous systems need to be resilient to these challenges to enable true long-term autonomy. Additionally, we present various logistical difficulties that we encountered in the context of our deployments in the Montmorency forest. The rest of this section will summarize the major lessons learned throughout the realization of this field report.

*Any static equipment slowly sinks in snow.* In the beginning of our deployment, we installed various static equipment to record position reference. A fixed GNSS reference receiver was installed to enable the RTK-GNSS and three *Trimble* S7 robotic total stations were positioned to measure ground truth robot poses in the *Garage* area, as shown in Figure 19a and Figure 19b. Since the precision of these systems relies on the fact that the equipment remains static, we invested a considerable effort in shoveling and stabilizing snow in the areas where this equipment was installed to prevent the sinkage (see Figure 19a). However, despite our efforts, we observed that the tripods carrying the equipment had sunk in the snow by several centimeters. The movement caused by this sinking effect made the affected ground truth measurements unusable. The outside temperature oscillating around the water freezing point and the snow and rain precipitations shown in Figure 9 led to snow melting, further increasing the tripod sinkage. Thus, significant effort and potentially using fixed structures is required to use such static equipment in winter conditions.

*Deep snow significantly affects UGV mobility.* As highlighted by Stansbury et al. (2004) and Lever et al. (2013), vehicle immobilization is a significant hazard when navigating on snow-covered terrain. We have attempted to manually drive the Warthog on a deep snow cover (i.e., 0.7 m), as shown in Figure 19c. During these runs, we noticed that robot mobility was significantly affected, especially during turning maneuvers. Also, the UGV chassis would float in sufficiently deep snow, resulting in reduced traction. At this point, turning maneuvers would make the robot sink even deeper, eventually requiring a human intervention to recover the robot. To prevent immobilization, backtracking maneuvers were necessary to compact the snow under the robot to maintain a minimal level of traction. This experience shows that immobilization prevention and recovery are required for mobile robot operation spanning multiple seasons in forest environments.

*Winter weather significantly affected the battery capacity of the Warthog UGV.* Outside temperatures during this deployment reached a minimum of $-15.5\,°C$, as shown in Figure 9. In our configuration, the Warthog is equipped with lead-acid batteries, whose capacity is known to be affected by low temperatures. Additionally, the CAMSO ATV T4S tracks that the robot used to drive on snow-covered terrain significantly increased energy consumption compared to using wheels on solid ground. These factors contributed to a UGV battery depletion that occurred during run R7, as captured by the photo in Figure 19d. The robot recovery required a gas generator to charge the batteries, and the operators were required to drive back to the *Garage* POI in the middle of the night. Thus, assuming that the vehicle battery capacity is stable for a period spanning multiple seasons is false and will eventually lead to a system power outage. Autonomous vehicle traversal planning should be conducted conservatively because the system recovery in remote environments is a costly operation. Future planners should include safety margins to take autonomy loss into account for mobile robots deployed in subzero weather.

*Resiliency to illumination variance is key to deploying mobile robots in northern environments.* Day length is subject to high variation in higher latitudes. To highlight this phenomenon, Figure 20 compares the sun radiation measured in the Montmorency forest during the deployment week and during the summer and winter solstice weeks. The days of the winter solstice week are short and receive minimum daylight. We argue that systems should be resilient to low- or no-illumination conditions to enable year-long autonomy in the northern environments. In remote areas, no artificial light sources such as streetlights or buildings are present to provide illumination for the robot. As discussed by Congram and Barfoot (2021), vision-based localization is affected by low illumination conditions, even when using headlights. However, making robots resilient to navigate in dark environments is key to enable autonomy in the discussed environment. As highlighted by Krüsi et al. (2015), lidar-based localization is resilient to high illumination variations, whereas vision-based approaches may fail (Paton et al., 2017). This is due to the fact that lidars are active sensors that do not rely on an external energy source to produce measurements. Moreover, the lack of daylight has another practical impact. Deploying mobile robots in night conditions is difficult due to low visibility, low temperatures, and operator fatigue, as can be seen in the system recovery snapshot shown in Figure 19d.

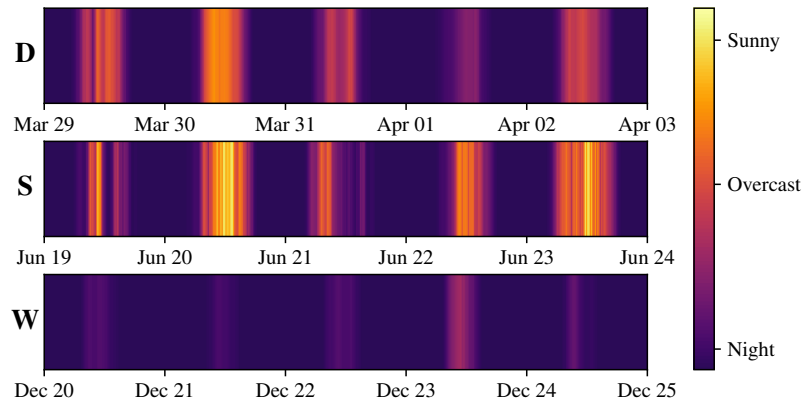

**Figure 20.** Sun radiation measured at the Montmorency forest for three distinct weeks. On top is the week during which this deployment (**D**) was taking place. In the middle is the week containing the 2021 summer (**S**) solstice. At the bottom is the week containing the 2020 winter (**W**) solstice. The coloring is proportional to the measured sun radiation every hour.

*Resilience to multiseasonal change and unexpected events.* In a later deployment during the fall of 2021, we attempted to repeat all paths by using the maps shown in Figure 10 as reference. We successfully repeated paths *A* and *C* by localizing in a map that was recorded 209 days before. However, we had similar initialization problems as the ones discussed in Section 6.2 when attempting path *B*. After multiple trials, we managed to start a repeat run of *B*, but the robot became immobilized in a mud pit, as shown in Figure 21a. The repeat phase was canceled at this point, and the robot was manually driven to the *Garage* POI. This experience supports our insight that immobilization prevention, recovery, and dynamic map adjustment are key to enable multiseasonal navigation. Later, when attempting to repeat path *C*, we noticed that concrete blocks had been added to the path by the forest management (see Figure 21b). The robot was manually driven to avoid collision with these blocks and left to autonomously repeat the rest of the path. Once again, we argue that, to deploy a robotic system in a boreal forest for a period spanning multiple years, autonomous navigation should be adaptive to the significant changes in the environment encountered during those deployments. An interesting avenue for future work would be to use a global planning algorithm, similar to the one presented by Guo and Barfoot (2019) that takes multiseasonal change into account. Such a system could replan paths based on major dynamic events happening on a global path network.

## 7. Conclusion

This paper is a field report presenting over 18 km of autonomous path repeating in boreal forest trails and paths, including runs under harsh weather and high illumination variations. We have

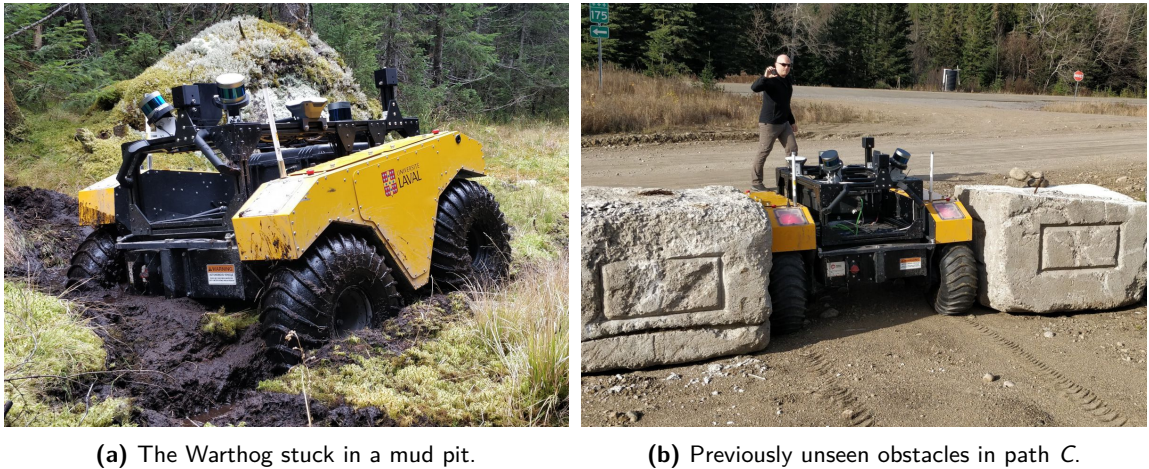

**(a)** The Warthog stuck in a mud pit.  **(b)** Previously unseen obstacles in path *C*.

**Figure 21.** Issues related to multiseason path repeating. (a) The Warthog stuck in a mud pit that was undetected during the teach phase of path *B*. Traversable terrain varies between seasons. (b) Obstacles that were moved in path *C* by human operators during summer. The presence of these obstacles required a human to take control of the UGV to avoid a collision.

described the WILN system, designed to enable wintertime lidar-based navigation, and deployed it in the Montmorency boreal forest to evaluate the performance of lidar-based and GNSS localization. We have highlighted the impact of the boreal forest biome and winter conditions on autonomous navigation technologies. Based on the data recorded during these runs, we have shown that forest trails are GNSS-denied environments and that localization uncertainty is high in such trails due to low geometrical constraints in the vehicle's longitudinal direction. Moreover, we have highlighted the impact of snow accumulation and dynamic changes in the environment on lidar-based localization over multiple days in wintertime. We have discussed the requirement to improve the adaptiveness of autonomous navigation to enable multiyear robotic deployments in boreal forests.

While we have shown that our WILN system is able to autonomously repeat paths through a boreal forest in harsh winter conditions, more work remains to enable true long-term autonomy. Lidar-based localization adaptive to changes in the environment is key to allowing mobile robot deployment in boreal forests throughout an entire year. The ability to detect large variations in traction conditions will be key to preventing system failures from vehicle immobilization. Improving interaction between the localization and control systems would also be beneficial to the system adaptability. For example, modulating controller commands based on the localization uncertainty could prevent localization and mapping failures. Furthermore, we have observed through data postprocessing that the reference maps built through the teach phase were close to global consistency. In the future, we would like to improve the mapping system to create globally consistent maps and enable localization initialization at any place in the map. This would in turn allow repeating kilometer-scale loops without requiring the UGV to stop for re-initialization at the end of each loop. Another interesting feature would be to enable global localization within the map. Such a feature would enable to solve the robot kidnapping problem and reset the robot localization at any point in the reference path.

## Acknowledgments

This research was supported by the Natural Sciences and Engineering Research Council of Canada (NSERC) through the grant CRDPJ 527642-18 SNOW (Self-driving Navigation Optimized for Winter) and FORAC. Support from the Montmorency forest staff was crucial to allow safe UGV deployment in a closed environment. Meteorological data were a courtesy of the Hydrometeorology

## ORCID

Dominic Baril https://orcid.org/0000-0002-7283-8406
Simon-Pierre Deschênes https://orcid.org/0000-0003-0594-5487
Olivier Gamache https://orcid.org/0000-0003-3278-5650
Maxime Vaidis https://orcid.org/0000-0002-1749-7207
Damien LaRocque https://orcid.org/0000-0003-3256-046X
Johann Laconte https://orcid.org/0000-0002-2274-9835
Vladimír Kubelka https://orcid.org/0000-0001-8393-9969
Philippe Giguère https://orcid.org/0000-0002-7520-8290
François Pomerleau https://orcid.org/0000-0003-1288-2744

Kilometer-scale autonomous navigation in subarctic forests: challenges and lessons learned · 1659Nieto, J., Guivant, J., Nebot, E., and Thrun, S. (2003). Real time data association for FastSLAM. In *2003 IEEE International Conference on Robotics and Automation*, volume 1, pages 412–418. IEEE.

Oliveira, L. F. P., Moreira, A. P., and Silva, M. F. (2021). Advances in forest robotics: A state-of-the-art survey. *Robotics*, 10(2).

Ostafew, C. J., Schoellig, A. P., and Barfoot, T. D. (2016). Robust constrained learning-based NMPC enabling reliable mobile robot path tracking. *International Journal of Robotics Research*, 35(13):1547–1563.

Paton, M., MacTavish, K., Ostafew, C. J., and Barfoot, T. D. (2015). It's not easy seeing green: Lighting-resistant stereo visual teach & repeat using color-constant images. *Proceedings - IEEE International Conference on Robotics and Automation*, 2015-June(June):1519–1526.

Paton, M., MacTavish, K., Warren, M., and Barfoot, T. D. (2016). Bridging the appearance gap: Multi-experience localization for long-term visual teach and repeat. In *2016 IEEE/RSJ International Conference on Intelligent Robots and Systems (IROS)*, volume 2016-November, pages 1918–1925. IEEE.

Paton, M., Pomerleau, F., MacTavish, K., Ostafew, C. J., and Barfoot, T. D. (2017). Expanding the limits of vision-based localization for long-term route-following autonomy. *Journal of Field Robotics*, 34(1):98–122.

Paton, M., MacTavish, K., Berczi, L.-P., van Es, S. K., and Barfoot, T. D. (2018). I can see for miles and miles: An extended field test of visual teach and repeat 2.0. In Hutter, M. and Siegwart, R., editors, *Field and Service Robotics*, pages 415–431, Springer International.

Pomerleau, F., Magnenat, S., Colas, F., Ming Liu, and Siegwart, R. (2011). Tracking a depth camera: Parameter exploration for fast ICP. In *2011 IEEE/RSJ International Conference on Intelligent Robots and Systems*, pages 3824–3829. IEEE.

Pomerleau, F., Colas, F., Siegwart, R., and Magnenat, S. (2013). Comparing ICP variants on real-world data sets. *Autonomous Robots*, 34(3):133–148.

Pomerleau, F., Krusi, P., Colas, F., Furgale, P., and Siegwart, R. (2014). Long-term 3D map maintenance in dynamic environments. In *2014 IEEE International Conference on Robotics and Automation (ICRA)*, number 1, pages 3712–3719. IEEE.

Pomerleau, F., Colas, F., and Siegwart, R. (2015). *A Review of Point Cloud Registration Algorithms for Mobile Robotics*. Now Publishers.

Ray, L., Jordan, M., Arcone, S. A., Kaluzienski, L. M., Walker, B., Koons, P. O., Lever, J., and Hamilton, G. (2020). Velocity field in the McMurdo shear zone from annual ground penetrating radar imaging and crevasse matching. *Cold Regions Science and Technology*, 173(February):103023.

Ren, R., Fu, H., Xue, H., Li, X., Hu, X., and Wu, M. (2021). LiDAR-based robust localization for field autonomous vehicles in off-road environments. *Journal of Field Robotics*, 38(8):1059–1077.

Russell, E. and Ritchie, J. C. (1988). Postglacial vegetation of Canada. *Bulletin of the Torrey Botanical Club*, 115:231.

Schall, O., Belyaev, A., and Seidel, H. P. (2005). Robust filtering of noisy scattered point data. *Point-Based Graphics, 2005 - Eurographics/IEEE VGTC Symposium Proceedings*, pages 71–77.

Schall, O., Belyaev, A., and Seidel, H. P. (2008). Adaptive feature-preserving non-local denoising of static and time-varying range data. *CAD Computer Aided Design*, 40(6):701–707.

Simon, P. (2015). Military robotics: Latest trends and spatial grasp solutions. *International Journal of Advanced Research in Artificial Intelligence*, 4.

Sprunk, C., Tipaldi, G. D., Cherubini, A., and Burgard, W. (2013). Lidar-based teach-and-repeat of mobile robot trajectories. *IEEE International Conference on Intelligent Robots and Systems*, pages 3144–3149.

Stansbury, R. S., Akers, E. L., Harmon, H. P., and Agah, A. (2004). Survivability, mobility, and functionality of a rover for radars in polar regions. *International Journal of Control, Automation and Systems*, 2(3):343–353.

Takasu, T. and Yasuda, A. (2009). Development of the low-cost RTK-GPS receiver with an open source program package RTKLIB. *International Symposium on GPS/GNSS*, (October):4–6.

Vaidis, M., Giguere, P., Pomerleau, F., and Kubelka, V. (2021). Accurate outdoor ground truth based on total stations. *Proceedings - 2021 18th Conference on Robots and Vision, CRV 2021*, pages 1–8.

Van Brummelen, J., O'Brien, M., Gruyer, D., and Najjaran, H. (2018). Autonomous vehicle perception: The technology of today and tomorrow. *Transportation Research Part C: Emerging Technologies*, 89(July 2017):384–406.

Williams, S. and Howard, A. M. (2009). Developing monocular visual pose estimation for arctic environments. *Journal of Field Robotics*, 27(2):145–157.
Field Robotics, July, 2022 · 2:1628–1660